\documentclass[twocolumn,showpacs,amsmath,amssymb,aps,prx,footinbib,floatfix,superscriptaddress,longbibliography, nofootinbib]{revtex4-2}
\usepackage{graphicx}
\usepackage{braket}
\usepackage{savesym}
\usepackage{amsfonts}
\usepackage{bm}
\usepackage{color}
\usepackage{subfigure}
\usepackage{wasysym}
\usepackage{upgreek}
\usepackage{float}
\usepackage{dsfont}
\usepackage{mathtools}
\usepackage[usenames,dvipsnames,svgnames,table]{xcolor}
\usepackage{hyperref}
\usepackage{hhline} 
\usepackage{multirow} 
\usepackage[normalem]{ulem}

\usepackage{booktabs}
\usepackage{placeins}

\makeatletter
\let\newfloat\newfloat@ltx
\makeatother
\usepackage{algorithm}
\usepackage{algorithmic}

\usepackage{makecell}
\usepackage[english]{babel}

\usepackage{format}
\usepackage{utfsym}

\begin{document}

\title{Training Multi-layer Neural Networks
       on Ising Machine}

\author{Xujie Song}
\affiliation{School of Vehicle and Mobility,  Tsinghua University,  Beijing,  China}

\author{Tong Liu}
\affiliation{School of Vehicle and Mobility,  Tsinghua University,  Beijing,  China}

\author{Shengbo Eben Li}
\thanks{Corresponding author.}
\affiliation{School of Vehicle and Mobility,  Tsinghua University,  Beijing,  China}

\author{Jingliang Duan}
\affiliation{School of Vehicle and Mobility,  Tsinghua University,  Beijing,  China}

\author{Wenxuan Wang}
\affiliation{School of Vehicle and Mobility,  Tsinghua University,  Beijing,  China}

\author{Keqiang Li}
\affiliation{School of Vehicle and Mobility,  Tsinghua University,  Beijing,  China}

\date{November 6, 2023}

\begin{abstract}
As a dedicated quantum device, Ising machines could solve large-scale binary optimization problems in 
milliseconds.
There is emerging interest in utilizing Ising machines to train feedforward neural networks due to the prosperity of generative artificial intelligence.
However, existing methods can only train single-layer feedforward networks because they can not handle complex nonlinear network topology.
This paper proposes an Ising learning algorithm to train quantized neural network (QNN),
where two essential techinques are incorporated, namely binary representation of topological network and order reduction of loss function.
As far as we know, this is the first algorithm that can train multi-layer feedforward networks on Ising machines,
which provides an alternative to gradient-based backpropagation method.
Firstly, training QNN is formulated as a quadratic constrained binary optimization (QCBO) problem by representing the neuron connection and activation function as equality constraints.
All quantized variables are encoded by binary bits based on binary encoding protocol.
Secondly, QCBO is converted into a quadratic unconstrained binary optimization (QUBO) problem, which can be efficiently solved on Ising machines.
The conversion leverages both penalty function method and Rosenberg order reduction method,
which together eliminate the equality constraints and reduce high-order loss function into a quadratic one.
With some reasonable assumptions,
theoretical analysis shows that the space complexity of our algorithm is $\mathcal{O}(H^2L + HLN\log H)$, which quantifies the required number of spins on an Ising machine.
Finally, the effectiveness of this algorithm is validated with a simulated Ising machine using MNIST dataset.
After annealing 700 milliseconds,
the classification accuracy achieves 98.3\%.
Among 100 runs, the success probability of finding the optimal solution is 72\%.
Along with the increasing number of spins on Ising machine,
our algorithm has the potential to train deeper neural networks.
\end{abstract}

\maketitle

\section{Introduction}
Deep neural networks,
renowned for their formidable capacity to glean intricate patterns from extensive datasets,
stand as the fundamental cornerstones of artificial intelligence.
Today,
feedforward neural network
is the most popular network family because of its strong representation ability, fast inference speed, and flexible connection structure.
Typical examples include AlexNet \cite{krizhevsky2012imagenet}, ResNet \cite{he2016deep}, DenseNet \cite{huang2017densely}, and Transformer \cite{vaswani2017attention}.
However, the enormous computational expenditure in training neural networks is becoming a pivotal limit to generate better intelligence.
Ising machine, as a dedicated quantum device, has achieved a substantial number of qubits,
implying a potential to train neural networks in advanced paradigms.
The super-fast computing ability of large-scale Ising machine may greatly benefit the training of feedforward neural networks.

Stable and programable Ising machines have been developed based on various physical systems, including trapped ion, superconducting circuit, molecule, optical, optoelectronic, and electrical systems \cite{marandi2014network, inagaki2016large}.
Ising machines are particularly suitable for solving combinatorial optimization problems that are formatted as quadratic unconstrained binary optimization (QUBO) \cite{neukart2017traffic, king2021scaling}.
The general form of QUBO problem is mathematically described as $\min_\sigma \sum_{ij}Q_{ij}\sigma_i\sigma_j$,
where $\sigma_i,\sigma_j \in \{0,1\}$ denote the $i$-th and $j$-th binary variables,
and $Q_{ij}\in \mathbb{R}$ denotes the element of coefficient matrix.
Generally, classical computer faces significant computational challenge to solve QUBO problems,
due to the frequent presence of nonplanar couplings that make these problems NP-hard \cite{barahona1982computational}.
For Ising machines,
finding the optimal solution is equivalent to searching the ground state of an Ising system \cite{lucas2014ising}.
Therefore, a QUBO problem can be efficiently solved on Ising machines
because the convergence to ground state often occurs at super-high speed \cite{cen2022large}.
For example, a good enough solution of a max-cut problem with 100,000 binary variables can be obtained in 1 millisecond on coherent Ising machine \cite{honjo2021100}.
As a result, it would be revolutionary if the fast computing ability of Ising machines could be applied to train deep neural networks.

Thus far,
Ising machines have been used in several machine learning tasks,
such as the training of support vector machine \cite{willsch2020support},
boosting model \cite{neven2012qboost},
and clustering model \cite{kumar2018quantum}.
As for neural networks, it is known that Ising machines could be used to train Boltzmann machine \cite{korenkevych2016benchmarking, bohm2022noise, adachi2015application, niazi2023training} and dynamical energy network \cite{laydevant2023training, laydevant2021training}.
These two branches employ either a ultrafast statistical sampling or a local learning rule to approximate gradients.
In the first branch,
\citet{bohm2022noise} succeeded in training the restricted Boltzmann machine by ultrafast statistical sampling.
Restricted Boltzmann machine consists only a single-layer connection between the visible units and hidden units,
resulting in limited expression ability.
\citet{adachi2015application} adopted an unsupervised layer-by-layer training approach for training the deep belief network.
In each step, they trained a restricted Boltzmann machine composed of two adjacent layers in the network.
But the approach is not a true multi-layer training method because only one layer is trained at a time,
which is very possible to reach local optima rather than global optima.
\citet{niazi2023training} directly trained the entire deep Boltzmann machine on sparse Ising machines.
However, its training relies on Gibbs-sampling-based inference,
whose speed is slow when input dimension and network size are large.
The drawback makes training a deep Boltzmann machine impractical for large-scale dataset and real-time application \cite{salakhutdinov2010efficient}.
In the other branch,
\citet{laydevant2023training} trained a dynamical energy network by equilibrium propagation method,
where an Ising machine is used to compute steady neuron states for approximating gradients.
There are only undirected graphic connections in dynamical energy network,
in contrast to the directed graphic connections in feedforward network.
As a result, its inference is based on the convergence of neuron states according to an energy function,
rather than a sequential computation in layers.
Furthermore, each neuron state is represented by one spin limited as $\pm 1$,
unable to extend to higher precision value.

All aforementioned networks do not belong to today's popular feedforward neural networks.
In a feedforward neural network, the information flow is uni-directional,
from the input layer, through the hidden layers and to the output layer,
without any cycles or loops.
Previously,
only single-layer feedforward networks without activation function can be trained on Ising machines.
The single-layer network, if we do not consider activation function, is actually a linear regression problem,
and therefore its parameter identification can be easily converted into a QUBO problem \cite{date2021qubo}.
As far as we know,
we have not found any method that is able to handle multi-layer feedforward networks on standard Ising machines.
The challenge of training multi-layer network is mainly due to the complex network topology,
including directed graphic connection among layers and nonliear activation function (e.g. piecewise linear type).
This complexity further leads to high-order loss function,
which can not be dealt with on Ising machines.

This paper for the first time proposes an Ising learning algorithm that makes it feasible to train multi-layer feedforward networks on Ising machines.
This algorithm maps a network training problem into a QUBO problem by two essential techniques, namely binary representation of topological network and order reduction of loss function.
In the first step, the training of quantized neural network (QNN) is converted into a quadratic constrained binary optimization (QCBO) problem.
Two techniques are included in this step, which are the constraint representation of network topology and the binary representation of variables.
The constraint representation captures the feedforward topology of QNN by describing the linear transformation and activation function as equality constraints.
The binary representation constructs all optimizing variables based on the binary encoding protocal for decimal numbers, which builds the relationship between optimizing variables and Ising spins.
The second step further convert QCBO into a QUBO problem,
which contains two techniques: penalty function method and Rosenberg order reduction method.
The penalty function is used to eliminate all equality constraints,
yielding a high-order loss function. 
The Rosenberg order reduction is used to transform the high-order loss into a quadratic loss.
The quadratic loss function composed of only binary variables builds a standard QUBO problem suitable for Ising machine computation.
Theoretical analysis has shown that,
for a network with constant-width hidden layers,
the space complexity of the Ising learning algorithm is $\mathcal{O}(H^2L + HLN\log H)$,
which quantifies the required number of Ising spins.
Here,
$N$ is dataset size,
$L$ is network depth,
and $H$ is network width.
As a non-gradient training approach,
Ising learning algorithm unlocks a new paradigm to train deep neural networks.
In the future,
better artificial intelligence will be generated with the increasing number of spins on Ising machines.

The rest of this paper is organized as follows:
Section \ref{sec:method} describes the proposed Ising learning algorithm,
including the overall diagram, technical details, and space complexity analysis.
Section \ref{sec:result} demonstrates its correctness and feasibility by verifications on simulated Ising machine.
Section \ref{sec:discuss} discusses the algorithm applicability and future prospects.

\section{Methods}
\label{sec:method}

\begin{figure*}[htbp]%
    \centering
    \includegraphics[width=0.99\textwidth]{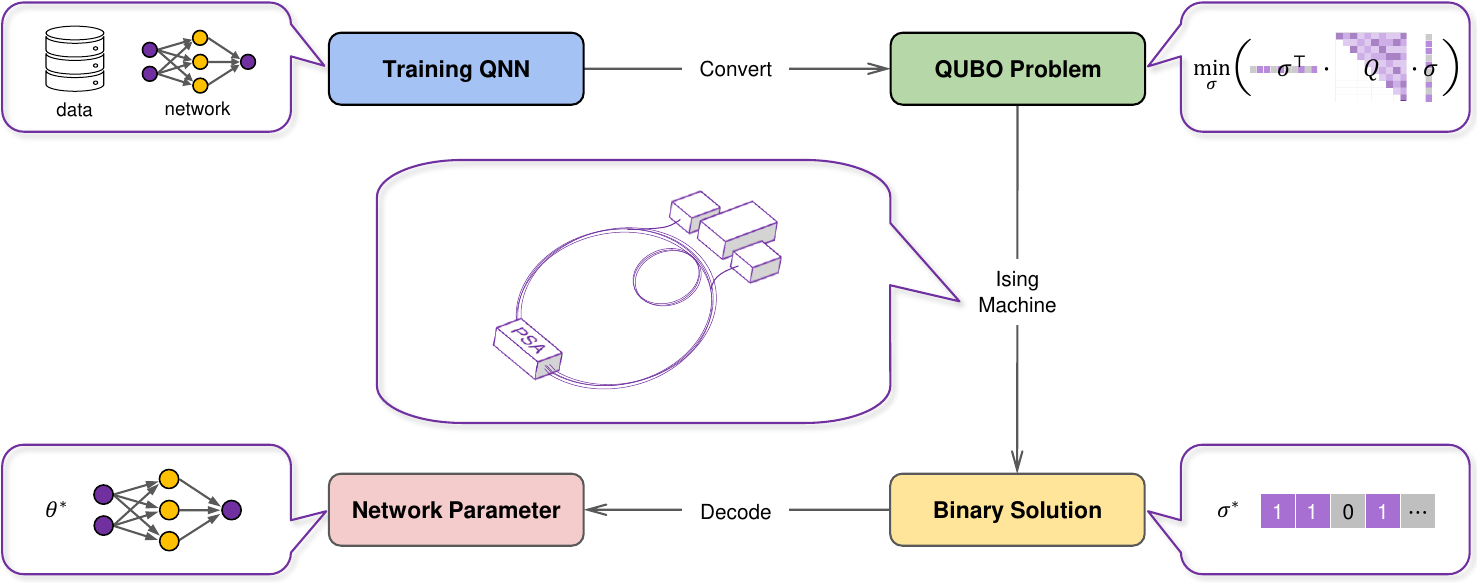}
    \caption{\textbf{Training workflow of Ising learning algorithm.}
             Firstly, training QNN is converted into a QUBO problem,
             whose coefficient matrix is $Q$.
             Then, the solution $\sigma^*$,
             which is a string of binary values,
             is solved by Ising machines.
             Finally, the network parameter $\theta^*$ is decoded from $\sigma^*$ based on binary encoding protocol.
             }
    \label{fig:workflow}
\end{figure*}

\subsection{Overall Diagram of Ising Learning}

\begin{figure*}[htbp]%
    \vspace{0.15in}
    \centering
     \includegraphics[width=0.9\textwidth]{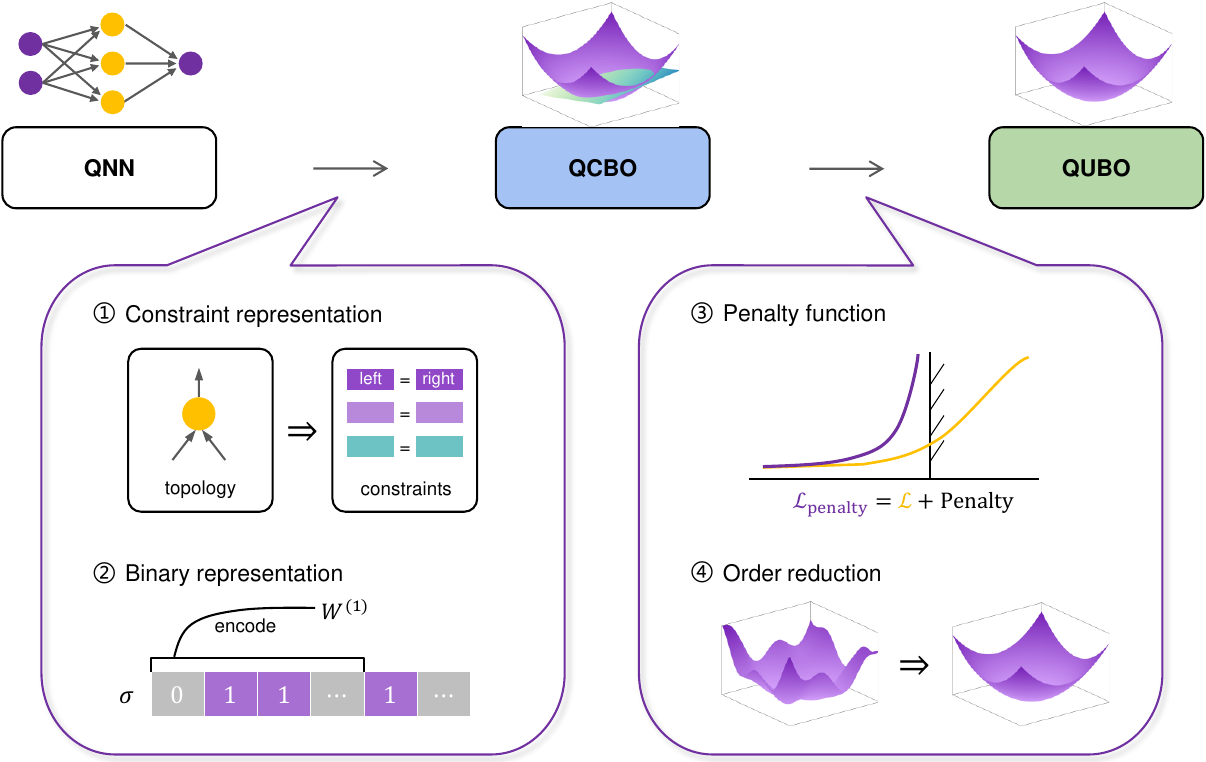}
    \caption{\textbf{Conversion from QNN to QUBO.}
             The training problem of QNN is initially described as a QCBO problem by constraint representation of network topology and binary representation of variables.
             The QCBO is then converted into a QUBO problem by penalty function method and Rosenberg order reduction method.
             }
    \label{fig:problem_transformation}
\end{figure*}

The proposed Ising learning algorithm focuses on a specialized feedforward neural network, i.e. quantized neural network (QNN).
A supervised learning task is built with mean squared error (MSE) loss function,
whose quantized parameters are trained with Ising machines.
The training problem for multi-layer QNN is described as Problem \ref{problem:supervised_learning}.
Notably, Ising machine exhibits high potential to suit diverse learning paradigms,
including self-supervised learning, reinforcement learning \cite{rlbook, duan2021distributional, guan2021direct, li2023miscellaneous}, etc.
\begin{problem}[]
    \label{problem:supervised_learning}
    Suppose that $f$ is a multi-layer network with quantized parameter $\theta$.
    The training problem is
    \begin{align}
        \theta^* = \arg \min_\theta \frac{1}{N} \sum_{i=1}^N \left(y_i - f\left( x_i \right)\right)^2,
        \label{eq:supervised_loss}
    \end{align}
    where $N$ is the size of dataset $\mathcal{D}$,
    and $(x_i, y_i)$ is the $i$-th data sample.
\end{problem}
The predicted output of network $f$ is computed as
$$f(x)=f^{(L)} \cdots f^{(2)} \circ f^{(1)}(x),$$
where $f^{(k)} (x) = g\left( W^{(k)} x + b^{(k)} \right)$ is the mapping in the $k$-th layer,
and $\circ$ means function composition.
This mapping contains a linear transformation with $W^{(k)}$ and $b^{(k)}$ as its weights,
and a nonlinear activation function $g$,
such as \textit{sign} and \textit{ReLU} \cite{rlbook}.
Therefore, $f$ has a complex feedforward topology,
and (\ref{eq:supervised_loss}) is a high-order nonlinear loss function
rather than a quadratic loss function with respect to $\theta$,
making it unsuitable for direct optimization on Ising machines.
To address this challenge,
we propose Ising learning algorithm,
whose training workflow is shown in Figure \ref{fig:workflow}.
In our algorithm, Problem \ref{problem:supervised_learning} is converted into a QUBO problem by two steps,
which is shown in Figure \ref{fig:problem_transformation}.

Firstly, Problem \ref{problem:supervised_learning} is rewrited as a QCBO problem.
There are two techniques in this step, including constraint representation of network topology and binary representation of variables.
The former technique expresses the feedforward topology, including linear transformation and activation function, as a series of equality constraints.
Specifically, each activation function is formulated as several polynomial equality constraints to avoid the appearance of nonlinear non-polynomial terms in loss function.
The latter technique encodes all decision variables by binary bits based on the binary encoding protocol for decimal numbers.
Each bit corresponds to one spin on Ising machine,
therefore it creates a clear linkage between decision variables and Ising spins.
In this way, training QNN is constructed as a QCBO problem.

Secondly,
QCBO is converted into a QUBO problem that is solvable on Ising machines.
The conversion process involves two techniques: penalty function method and Rosenberg order reduction method.
The former technique eliminates equality constraints by adding the squares of constraint function into loss function.
The resulting high-order loss function is then reduced to a quadratic one by iteratively applying Rosenberg order reduction,
which is the latter technique.
In each iteration,
one second-order factor in the loss function is substituted with one auxiliary binary variable, and then a Rosenberg polynomial is added into the loss function as a new positive penalty.
The two techniques together yield a quadratic loss function whose coefficients compose the elements of matrix $Q$ in a QUBO problem.

The conversion from Problem \ref{problem:supervised_learning} to QUBO  should be executed on classical computers,
whose output is the coefficient matrix $Q$.
The computation relies on center processing unit (CPU).
Subsequently,
QUBO is directly solved on an Ising machine,
which is a kind of quantum processing unit (QPU).
Its solution $\sigma^*$ is a string of 0-1 values,
which will be decoded back to the optimal network parameter $\theta^*$.
The overall workflow of Ising learning is shown in Algorithm \ref{alg:IL} and Figure \ref{fig:workflow}.
The computational burden of problem conversion and parameter decoding is generally low even for large-scale problems,
therefore our Ising learning algorithm can be very computationally efficient.

\begin{algorithm}[htbp]
    \floatname{algorithm}{Algorithm }
    \renewcommand{\algorithmicrequire}{\textbf{Input:}}
    \renewcommand{\algorithmicensure}{\textbf{Output:}}
    \caption{:\quad Ising learning}
    \label{alg:IL}
 \begin{algorithmic}[1]
    \vskip 0.05in
    \REQUIRE dataset $\mathcal{D}$.
    \vskip 0.1in
    \STATE Convert Problem \ref{problem:supervised_learning} into a QCBO problem.
    \vskip 0.05in
    \STATE Convert QCBO into a QUBO problem.\hfill \hfill $\triangleright$\hspace{0.049in}CPU
    \vskip 0.05in
    \STATE Solve QUBO on Ising machine\\
    \vskip 0.05in \hskip 0.8in
    $\sigma^* = \arg \min_\sigma\ \mathcal{L}_{\rm QUBO}$. \hfill $\triangleright$\ QPU
    \vskip 0.05in
    \STATE Decode parameter\\
    \vskip 0.05in \hskip 0.815in
    $\theta^* = \textit{decode}(\sigma^*)$.\hfill $\triangleright$\hspace{0.049in}CPU
    \vskip 0.1in
    \ENSURE optimal network parameter $\theta^*$.
 \end{algorithmic}
\end{algorithm}

\subsection{Convert Training QNN into QCBO Problem}
\label{sec:QCBO}

This section describes how to formulate training QNN as a combinatorial optimization problem in the QCBO format.
The loss function to be minimized is chosen as mean squared error (MSE):
\begin{align}
    \mathcal{L}_{\rm MSE} = \frac{1}{N} \sum_{i=1}^N (y_i - \hat{y}_i)^2,
    \label{eq:MSE}
\end{align}
where $\hat{y}_i$ is a decision variable representing predicted output.
In this context, $\hat{y}_i$ is not the inference output $f(x_i)$, but a decision variable representing all possible output.
When minimizing $\mathcal{L}_{\rm MSE}$,
we need to ensure that $\hat{y}_i$ is equal to the actual inference output $f(x_i)$ produced by QNN.
Therefore, certain constraints must be set up to capture the feedforward topology of the network.

\begin{figure*}[htbp]%
    \centering
    \includegraphics[width=0.5\textwidth]{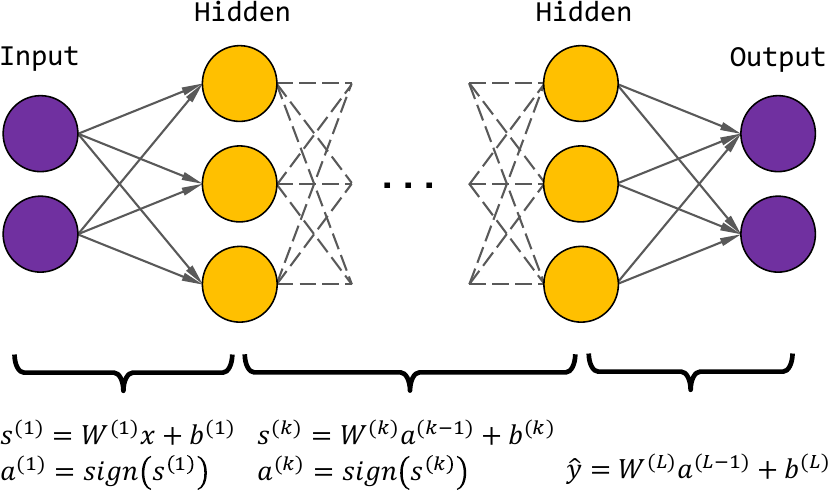}
    \caption{\textbf{Structure of QNN.}
             The $W^{(k)}$ and $b^{(k)}$ are quantized weight and bias of layer $k$, respectively.
             The \textit{sign} function is used as activation function except for the last layer.
             The pre-activation and post-activation value are represented as $s^{(k)}$ and $a^{(k)}$, respectively.
             The $x$ and $\hat y$ are quantized input and predicted output, respectively.
             }
    \label{fig:qnn}
\end{figure*}

For simplicity,
QNN in our paper is chosen as a fully-connected multi-layer network with constant-width hidden layers,
as shown in Figure \ref{fig:qnn}.
The \textit{sign} function is used as activation function,
which is a common choice in QNN \cite{hubara2016binarized}.
Table \ref{tab:qnn_variable} provides the summary of symbols used in the representation of network structure.

\begin{table}[htbp]
    \caption{\textbf{Symbols in quantized neural network.}}
    \label{tab:qnn_variable}
    \begin{center}
    \begin{small}
    \begin{tabular}{c|l}
        \toprule
        Symbol & Description\\
        \midrule
        $L$ & total number of layers\\
        $H$ & number of neurons in each hidden layer\\
        $n$ & number of neurons in the input layer\\
        $m$ & number of neurons in the output layer\\
        \hline
        $x$ & quantized input\\
        $\hat y$ & predicted output\\
        \hline
        $W^{(k)}$ & quantized weight in the $k$-th layer\\
        $b^{(k)}$ & quantized bias in the $k$-th layer\\
        \hline
        $s^{(k)}$ & pre-activation value in the $k$-th layer\\
        $a^{(k)}$ & post-activation value in the $k$-th layer\\
        \bottomrule
    \end{tabular}
    \end{small}
    \end{center}
\end{table}

As a standard feedforward neural network,
QNN is composed of a large amount of neurons.
There are two essential operations in each neuron: linear transformation and activation function.
To capture the network topology during optimization,
these two operations are viewed as equality constraints.
As for the linear transformation,
its equality constraint is
\begin{align}
    W^{(k)} a^{(k-1)} + b^{(k)} = s^{(k)}.
    \label{eq:linear_constraint}
\end{align}
As for the activation function,
its behavior is
\begin{align*}
    {\textit{sign}}\left(x\right) = 
    \begin{cases}
        +1,\ \ x\geq 0\\
        -1,\ \ x< 0
    \end{cases}.
\end{align*}
To capture this behavior,
we firstly come up with two constraints:
\begin{align}
    & a^{(k)} \odot s^{(k)} = r^{(k)}\label{sign_1}\\
    & a^{(k)} + 2r^{(k)} \geq 1,\label{sign_2}
\end{align}
where $a^{(k)} \in \{-1,+1\}^H$, $s^{(k)} \in \mathbb{Z}^H$,
operation $\odot$ represents element-wise multiplication,
and $r^{(k)} \in \mathbb{N}^H$ is an auxiliary variable.
The equality constraint (\ref{sign_1}) guarantees $a^{(k)}$ and $s^{(k)}$ have the same sign,
because $r^{(k)}$ is non-negative.
The value of $r^{(k)}$ will be equal to the absolute value of $s^{(k)}$ under constraint (\ref{sign_1}).
The inequality constraint (\ref{sign_2}) guarantees $a^{(k)}=+1$ when $s^{(k)}=0$.
Together, these two constraints guarantee $a^{(k)}=\textit{sign}\left(s^{(k)}\right)$.
In order to handle constraint (\ref{sign_2}) more conveniently in the following process,
an auxiliary variable $t^{(k)} \in \mathbb{N}^H$ is introduced to transfer the inequality constraint (\ref{sign_2}) into an equality constraint:
\begin{align}
    a^{(k)} + 2r^{(k)} = 1 + t^{(k)}.\label{sign_3}
\end{align}

The constraint (\ref{eq:linear_constraint}),
(\ref{sign_1}), and (\ref{sign_3}) need to be defined for all layers except the last layer,
while only constraint (\ref{eq:linear_constraint}) is needed in the last layer,
as this layer does not have activation function.
Furthermore, these equality constraints should be defined for all data samples.
For example, the linear transformation constraint (\ref{eq:linear_constraint}) for the $k$-th layer actually becomes:
\begin{align}
    W^{(k)} a^{(k-1)}_i + b^{(k)} = s^{(k)}_i,\ \forall i.    \label{eq:linear_constraint_sample}
\end{align}
In the following context,
variables with subscript $i$ will denote the $i$-th sample.
For example, $a^{(1)}_i$ denotes the post-activation value in the first layer when $x_i$ is sent into the network.

The set of decision variables is $\{W^{(k)}, b^{(k)}, s_i^{(k)}, r^{(k)}_i,$ $ t^{(k)}_i, a^{(k)}_i, \hat{y}_i\mid k=1,2,...,L\ \textit{and}\ i=1,2,...,N\}$,
which are needed to be optimized to minimize (\ref{eq:MSE}) with all mentioned constraints.
To map these variables on Ising spins,
we use the technique named binary representation of variables.
The variables are encoded by several bits based on the binary encoding protocol for decimal numbers.
The value domain of each bit is $\{0,1\}$,
corresponding to one spin on an Ising machine.
For example, the bias in the first layer is encoded by
\begin{align*}
    b^{(1)} = \sum_{j}2^{j}\cdot \sigma_{b}^{(1)}(j),
\end{align*}
where $\sigma_{b}^{(1)}(j)\in \{0,1\}^H$ is a binary variable.
We denote $\sigma$ as the set that contains all the binary variables used in encoding.
Through this binary representation technique,
$W^{(1)}$, $b^{(1)}$, and $s^{(1)}_i$ are all encoded as quantized values.
Then the constraint $W^{(1)} x_i + b^{(1)} = s^{(1)}_i$ requires that $x_i$ should also be quantized.
To meet this requirement,
we quantize $x_i$ as integer by rounding,
where $x_i\in \mathbb{X}^n$ and $\mathbb{X}=[-2^B,2^B]$.
Here, $B$ denotes the bit width of input data.
Typical classification and regression problems can be modified to this kind of dataset by introducing scale factor,
so there is no loss of generality.

So far, training QNN is formulated as a QCBO problem,
as described in Problem \ref{problem:qcbo}.
\begin{problem}[(QCBO)]
    \label{problem:qcbo}
    The training problem is
    \begin{align*}
        &\hskip 0.7in \sigma^*= \arg \min_\sigma\ \mathcal{L}_{\rm MSE}\\
        & \text{subject to}\\
        & \hskip 0.3in \text{constraints (\ref{eq:linear_constraint}), (\ref{sign_1}), and (\ref{sign_3}) for layer}\ k\quad\\
        & \hskip 0.3in \text{and sample}\ i,\ \forall k\ \forall i
    \end{align*}
    where $\mathcal{L}_{\rm MSE}$ is shown in (\ref{eq:MSE}).
\end{problem}

\subsection{Convert into Solvable QUBO Problem}
\label{sec:convert}
The Problem \ref{problem:qcbo} which is a QCBO problem can not be deployed on Ising machine.
It needs to be converted into a QUBO format,
where no constraints are involved.
The conversion from QCBO to QUBO needs two techniques:
(a) eliminating all constraints by penalty function method;
(b) reducing the order of loss function into quadratic order by Rosenberg polynomials.
The conversion process is summarized in Algorithm \ref{alg:conversion_scheme}.

\begin{algorithm}[htbp]
    \floatname{algorithm}{Algorithm }
    \renewcommand{\algorithmicrequire}{\textbf{Input:}}
    \renewcommand{\algorithmicensure}{\textbf{Output:}}
    \caption{:\\Conversion Process (QCBO $\xrightarrow{}$ QUBO)}
    \label{alg:conversion_scheme}
 \begin{algorithmic}[1]
    \vskip 0.05in
    \REQUIRE loss function $\mathcal{L}_{\rm MSE}$, constraints set $\Phi$ where the $j$-th constraint is $\phi_j=0$.
    \vskip 0.1in
    \STATE \textit{// penalty function method}
    \STATE $\mathcal{L}_{\rm penalty} \leftarrow \mathcal{L}_{\rm MSE}$
    \FOR{the $j$-th constraint $\phi_j=0$ in $\Phi$}
        \STATE \quad $\mathcal{L}_{\rm penalty} \leftarrow \mathcal{L}_{\rm penalty} + \rho\phi_j^2$
    \ENDFOR
    \vspace{0.1in}
    \STATE \textit{// order reduction method}
    \STATE $\mathcal{L}_{\rm QUBO} \leftarrow \mathcal{L}_{\rm penalty}$
    \STATE $\Omega =\{polynomial\ terms\ \omega\ in\ \mathcal{L}_{\rm QUBO} \mid \ \omega's\ order > 2\}$
    \STATE Count frequency of each factor $u_1u_2$ in $\Omega$
    \WHILE {the order of $\mathcal{L}_{\rm QUBO} > 2$}
        \STATE \quad Find the most frequent factor $u_1u_2$ in $\Omega$
        \STATE \quad Substitute $u_1u_2$ in $\mathcal{L}_{\rm QUBO}$ with $v$
        \STATE \quad $\mathcal{L}_{\rm QUBO} \leftarrow \mathcal{L}_{\rm QUBO} + \lambda h(u_1,u_2,v)$
        \STATE \quad Update $\Omega$ and factors' frequency
    \ENDWHILE
    \vskip 0.1in
    \ENSURE quadratic loss function $\mathcal{L}_{\rm QUBO}$.
 \end{algorithmic}
\end{algorithm}

In the first technique, we adopt the penalty function method \cite{bazaraa2013nonlinear} to eliminate all equality constraints.
The loss function (\ref{eq:MSE}) is reformulated as
\begin{align}
    \label{eq:big_M_loss}
    \mathcal{L}_{\rm penalty} &= \frac{1}{N} \sum_{i=1}^N (y_i - \hat{y}_i)^2 \nonumber\\
    &+ \rho \sum_{i=1}^N \left\Vert W^{(1)} x_i + b^{(1)} - s^{(1)}_i \right\Vert^2 \nonumber\\
    &+ \rho \sum_{i=1}^N \sum_{k=2}^{L-1} \left\Vert W^{(k)} a^{(k-1)}_i + b^{(k)} - s^{(k)}_i \right\Vert^2 \nonumber\\
    &+ \rho \sum_{i=1}^N \left\Vert W^{(L)} a^{(L-1)}_i + b^{(L)} - \hat{y}_i \right\Vert^2 \nonumber\\
    &+ \rho \sum_{i=1}^N \sum_{k=1}^{L-1} \left\Vert a^{(k)}_i \odot s^{(k)}_i - r^{(k)}_i \right\Vert^2 \nonumber\\
    &+ \rho \sum_{i=1}^N \sum_{k=1}^{L-1} \left\Vert a^{(k)}_i + 2r^{(k)}_i - 1 - t^{(k)}_i \right\Vert^2,
\end{align}
where $L$ is the total number of layers,
$\left\Vert\cdot\right\Vert$ is vector norm,
and $\rho$ is a positive constant.
The second to the forth rows in (\ref{eq:big_M_loss}) are the constraint penalties for linear transformation,
and the last two rows are the constraint penalties for activation function.
With large enough $\rho$,
the cost of violating constraints is greater than that of increasing prediction accuracy,
which benefits the satisfaction of equality constraints \cite{zaman2021pyqubo}.
The loss function (\ref{eq:big_M_loss}),
whose order is greater than two,
forms an high-order unconstrained binary optimization problem.

In the second technique, Rosenberg polynomial \cite{mandal2020compressed} is used to reduce the order of loss function (\ref{eq:big_M_loss}),
thereby obtaining an unconstrained optimization problem with quadratic loss function $\mathcal{L}_{\rm QUBO}$.
The general form of Rosenberg polynomial is
$$h(u_1,u_2,v) = 3v+u_1 u_2 - 2u_1 v -2u_2 v,$$
where $u_1, u_2, v \in \{0,1\}$.
There are two properties of Rosenberg polynomial:
\renewcommand\labelitemi{\small$\bullet$}
\begin{itemize}
    \item $h(u_1, u_2, v)\geq 0,\ \forall u_1, u_2, v$;
    \item $h(u_1, u_2, v) = 0$ if and only if $v=u_1u_2$.
\end{itemize}
These properties imply that
$h(u_1, u_2, v)=0$ when $v=u_1 u_2$,
and $h(u_1,u_2,v)>0$ when $v\neq u_1 u_2$.
It means that Rosenberg polynomial takes its minimum value if and only if $v=u_1 u_2$.

Any high-order binary optimization problem can be reduced into a quadratic binary optimization problem
by iteratively running the Rosenberg order reduction method.
In each iteration,
this method substitutes a multiplication of two binary variables $u_1u_2$ with one auxiliary binary variable $v$ and adds $h(u_1,u_2,v)$ into the loss function with a large positive coefficient.
The large coefficient drives $h(u_1,u_2,v)$ to reach its minimum value where $v=u_1 u_2$ holds.
As a result,
the second-order factor $u_1 u_2$ is equivalently replaced by a first-order factor $v$ in each iteration,
finally yielding a quadratic loss function.
An example is shown in Appendix \ref{appendix:rosenberg} to illustrate the process of order reduction.

After converting (\ref{eq:big_M_loss}) into a quadratic loss function $\mathcal{L}_{\rm QUBO}$,
Problem \ref{problem:qcbo} becomes a QUBO problem, as described in Problem \ref{problem:QUBO}.

\begin{problem}[(QUBO)]
    \label{problem:QUBO}
    The training problem is
    \begin{align}
        &\sigma^* = \arg \min_\sigma\ \mathcal{L}_{\rm QUBO},  \nonumber
    \end{align}
    where $\mathcal{L}_{\rm QUBO}$ is a quadratic loss function with $Q$ as its coefficient matrix.
\end{problem}

\subsection{Space Complexity Analysis}
\label{sec:space_complexity}
Our algorithm maps all optimizing variables on Ising spins,
therefore its space complexity is needed to be figured out.
The space complexity describes the increasing trend of spin number when the training problem becomes more complex.
Using fewer spins is advisable, as supported number of spins remains a significant bottleneck in today's Ising machines.

The space complexity depends on both algorithm design and network configuration,
including bit width of parameters, parameter freeze strategy, etc.
To analyze our algorithm without considering network type,
we take a specific configuration of QNN as an example (see Appendix \ref{appendix:configuration} for more details).
This configuration is also used in algorithm verification in Section \ref{sec:result}.

Before analysis,
we need to specify how many decimal decision variables are used and how many bits are used in each variable's encoding.
Table \ref{tab:num_spins} lists all the decision variables.
As for the number of bits,
it depends on one simple principle that
bit number should be consistent to the value range of decimal variable.
For example, constraint (\ref{eq:linear_constraint_sample}) requires that the value range of $s^{(k)}_i$ should cover that of $W^{(k)} a_i^{(k-1)} + b^{(k)}$ and we know that the value range of elements in $W^{(k)} a_i^{(k-1)} + b^{(k)}$ is $2H$,
therefore the elements in $s^{(k)}_i$ should contain $\lfloor \log_2 2H \rfloor + 1$ bits.
Based on this calculation principle,
Table \ref{tab:num_spins} lists the number of spins used in each variable.
By summing all spin numbers in Table \ref{tab:num_spins},
the space complexity is obtained:
\begin{align}
    \mathcal{O}(&nH + mN\log H + HN\log n + H^2 L \nonumber\\
    & + BHN + mH\log H  + HLN\log H).
    \label{eq:space_complexity}
\end{align}

\begin{table*}[htbp]
    \caption{\textbf{Number of spins used in the encoding of variables.}
             The left column shows the variable name.
             The central column shows the required number of spins to encode a single variable with a specific index.
             The right column shows the valid domain for superscript index and subscript index,
             which are corresponding to layer and sample, respectively.
             The total number of spins for one type of variable is equal to the multiplication of the range of index domain and the number of spins for a single variable.}
    \label{tab:num_spins}
    \begin{center}
    \begin{small}
    \begin{tabular}{c|c|l}
        \toprule
        Variable & Number of spins & Index domain\\
        \midrule
        $W^{(1)}$ & $nH$ & N/A\\
        $b^{(1)}$ & $H\lfloor \log_2 n2^{B+1}+1\rfloor$ & N/A\\
        $W^{(k)}$ & $H^2$ & $k=2,3,...,L-1$\\
        $W^{(L)}$ & $mH\lfloor \log_2 2H + 1\rfloor$ & N/A\\
        $b^{(L)}$ & $m\lfloor \log_2 2H + 1\rfloor$ & N/A\\
        \hline
        $s^{(1)}_i$ & $H\lfloor \log_2 n2^{B+2}+1\rfloor$ & $i=1,2,...,N$\\
        $s^{(k)}_i$ & $H\lfloor \log_2 2H+1\rfloor$ & $k=2,3,...,L-1$ and $i=1,2,...,N$\\
        $r^{(1)}_i$ & $H\lfloor \log_2 3n2^{B}+1\rfloor$ & $i=1,2,...,N$\\
        $r^{(k)}_i$ & $H\lfloor \log_2 2H+1\rfloor$ & $k=2,3,...,L-1$ and $i=1,2,...,N$\\
        $t^{(1)}_i$ & $H\lfloor \log_2 3n2^{B+1}+1\rfloor$ & $i=1,2,...,N$\\
        $a^{(k)}_i$ & $H$ & $k=1,2,...,L-1$ and $i=1,2,...,N$\\
        $\hat y_i$ & $m\lfloor \log_2 4H+1 \rfloor$ & $i=1,2,...,N$\\
        \bottomrule
    \end{tabular}
    \end{small}
    \end{center}
\end{table*}

As shown in (\ref{eq:space_complexity}),
the space complexity relates with network depth $L$, network width $H$,
dataset size $N$,
dimension of input features $n$, dimension of output features $m$,
and bit width of input features $B$.
Because $n, m$, and $B$ are properties of data sample,
which are always predefined values in general training tasks,
they can be assumed as constants.
As a result,
the space complexity can be simplified to be
\begin{align}
    \mathcal{O}(&H^2L + HLN\log H).
    \label{eq:space_complexity_simple}
\end{align}

The space complexity (\ref{eq:space_complexity_simple}) implies that
(a) the number of spins is linearly proportional to dataset size $N$;
(b) the number of spins is linearly proportional to network depth $L$;
and (c) the number of spins is quadratically proportional to network width $H$.
Usually, the dataset size $N$ is significantly bigger than network width $H$ and network depth $L$.
Therefore,
the bottleneck in space complexity is the size of dataset rather than the size of network.

\section{Results}
\label{sec:result}
\begin{figure}[htbp]%
    \centering
    \includegraphics[width=0.34\textwidth]{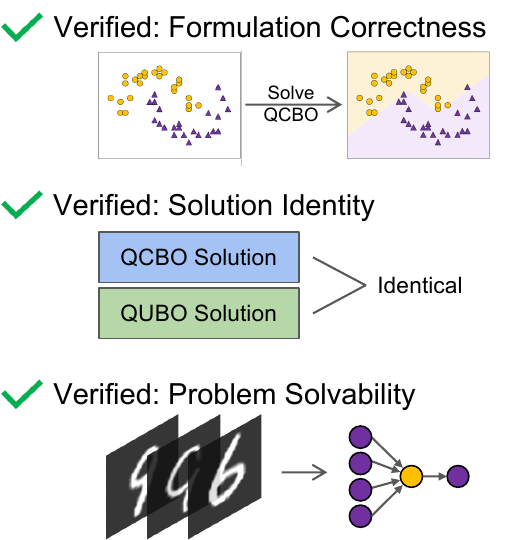}
    \caption{\textbf{Correctness verification of Ising learning algorithm.}
             The three verifications demonstrate the formulation correctness of QCBO problem, the solution identity between QCBO and QUBO problem, and the problem solvability with Ising machines.
             }
    \label{fig:verified_things}
\end{figure}

To demonstrate the correctness of Ising learning algorithm,
we implement three kinds of verification,
as shown in Figure \ref{fig:verified_things}.
They include formulation correctness,
solution identity,
and problem solvability,
which are respectively described in Section \ref{verification:formulation}, \ref{verification:identity}, and \ref{verification:solvability}.
The first verification aims to examine the formulation correctness of Problem \ref{problem:qcbo} (QCBO).
The second verification aims to examine the solutions identity of Problem \ref{problem:qcbo} (QCBO) and Problem \ref{problem:QUBO} (QUBO), thereby validating the correctness of conversion process in section \ref{sec:convert}.
The third verification aims to examine the solvability of Problem \ref{problem:QUBO} (QUBO) on typical Ising machines.
For simplicity,
we set some common configurations for QNNs used in our verification,
as shown in Appendix \ref{appendix:configuration}.

\subsection{Verification of Formulation Correctness}
\label{verification:formulation}
\begin{figure}[htbp]%
    \centering\hspace{-0.3in}
    \includegraphics[width=0.31\textwidth]{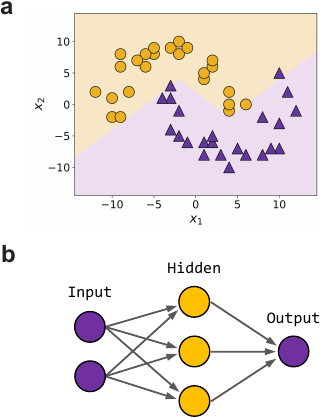}
    \caption{\textbf{Verification of formulation correctness.}
             (a) Data samples and training result:
             The point color implies data labels.
             $x_1$ and $x_2$ are two-dimensional input features.
             The area color implies decision boundary of the solved network,
             who correctly fits the given dataset.
             (b) Two-layer QNN:
             It contains 1 hidden layer with 3 hidden neurons.
             }
    \label{fig:moon_dataset}
    \vspace{-0.15in}
\end{figure}

Problem \ref{problem:qcbo} (QCBO) is a combinatorial optimization problem designed for training QNN.
Its formulation correctness can be verified by checking whether the solved network correctly fits the given dataset.
In this verification,
we use a two-moon dataset \cite{pmlr-v119-van-amersfoort20a},
which contains 50 samples with two kinds of labels.
As shown in Figure \ref{fig:moon_dataset}(a),
the data samples distribute in 2D space,
and their points entangle like two moons.
The network has 1 hidden layer with 3 hidden neurons,
as shown in Figure \ref{fig:moon_dataset}(b).

In the corresponding QCBO problem,
the number of binary variables is 3839.
Then, we solve the problem by Gurobi optimizer \cite{gurobi}, one of the best large-scale optimizer.
After running 15 seconds, the final loss achieves 0.31 and the classification accuracy achieves 98\%.
As shown in Figure \ref{fig:moon_dataset}(a),
the decision boundary of the solved network,
which is divided by two colored areas,
correctly fits the given dataset.
The result successfully demonstrates the formulation correctness of Problem \ref{problem:qcbo}.

\subsection{Verification of Solution Identity}
\label{verification:identity}
Section \ref{sec:convert} converts the QCBO problem into a QUBO problem.
In this section, three verification tasks are designed to check whether QUBO's solution is identical to QCBO's.
The solution identity is a strong evidence to illustrate the correctness of conversion process,
because the conversion is expected to eliminate constraints without changing problem's optimality.

\begin{figure*}[htbp]
    \centering
    \includegraphics[width=0.99\textwidth]{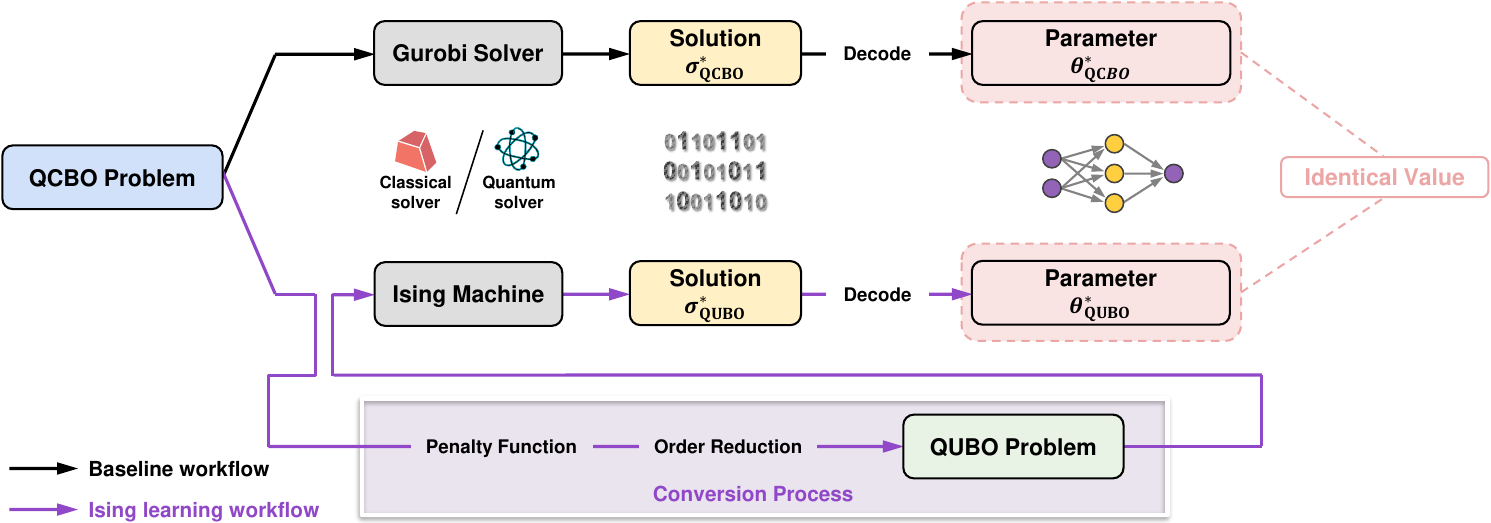}
    \caption{\textbf{Verification strategy for solution identity.}
             The conversion process in purple block converts QCBO (Problem \ref{problem:qcbo})
             to QUBO (Problem \ref{problem:QUBO}).
             Our aim is to validate the identity of $\theta^*_{\rm QCBO}$ and $\theta^*_{\rm QUBO}$, thereby verifying this conversion process.
             The solution $\sigma^*_{\rm QCBO}$ is solved by classical solvers, e.g. Gurobi optimizer.
             The solution $\sigma^*_{\rm QUBO}$ is solved by quantum solvers, e.g. Ising machines.
             The network parameters $\theta^*_{\rm QCBO}$ and $\theta^*_{\rm QUBO}$ are decoded from solutions.
             The black line indicates the baseline workflow for solving QCBO.
             The purple line indicates the Ising learning workflow.}
    \label{fig:verify_scheme}
\end{figure*}

\begin{figure*}[htbp]%
    \centering
    \includegraphics[width=0.8935\textwidth]{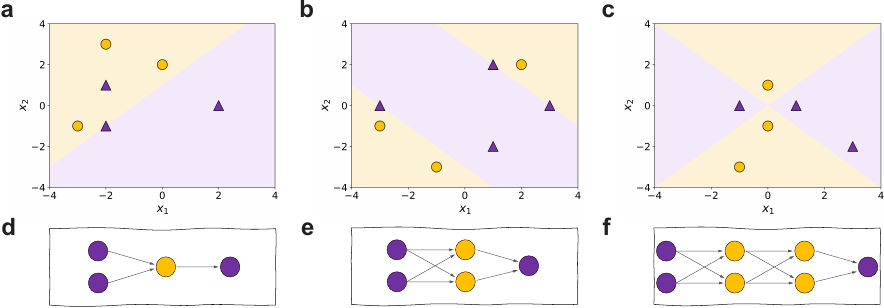}
    \caption{\textbf{Data samples, network structures and training results on three classification tasks.}
             The point color implies the data labels.
             The $x_1$ and $x_1$ are two-dimensional input features.
             We found the network decision boundaries of parameters $\theta^*_{\rm QCBO}$ and $\theta^*_{\rm QUBO}$ are the same in each task.
             The area color implies their network decision boundaries.
             (a) Dataset and training result on Task \uppercase\expandafter{\romannumeral1}.
             (b) Dataset and training result on Task \uppercase\expandafter{\romannumeral2}.
             (c) Dataset and training result on Task \uppercase\expandafter{\romannumeral3}.
             (d) Network used in Task \uppercase\expandafter{\romannumeral1}.
             Task \uppercase\expandafter{\romannumeral1} is an under-parameterized task because of insufficient network parameters.
             (e) Network used in Task \uppercase\expandafter{\romannumeral2}.
             Task \uppercase\expandafter{\romannumeral2} is a well-parameterized task.
             (f) Network used in Task \uppercase\expandafter{\romannumeral3}.
             Task \uppercase\expandafter{\romannumeral3} is also a well-parameterized task.
             }
    \label{fig:iden_sol}
\end{figure*}

The verification strategy is shown in Figure \ref{fig:verify_scheme}.
For QCBO,
we solve it by Gurobi.
For QUBO,
we solve it by tree decomposition method \cite{dwave},
due to the difficulty to access high-performance Ising machine with sufficient spin number and coefficient resolution.
To perform systematic evaluation,
one under-parameterized task (Task \uppercase\expandafter{\romannumeral1})
and two well-parameterized tasks (Task \uppercase\expandafter{\romannumeral2} and \uppercase\expandafter{\romannumeral3})
are selected.
Over-parameterized situation is not involved in the verification,
because there will be multiple solutions.
The data samples, network structures, and training results are all summarized in Figure \ref{fig:iden_sol}.
The number of used spins is listed in Table \ref{tab:spin_num}.
The loss and accuracy are listed in Table \ref{tab:indicator}.

\begin{table}[htbp]
    \setlength{\tabcolsep}{7pt}
    \vskip -0.15in
    \caption{\textbf{The number of binary variables used in each task.}
             The additional number of binary variables in QUBO compared to QCBO originates from the auxiliary variables in Rosenberg polynomial.}
    \label{tab:spin_num}
    \vskip 0.15in
    \begin{center}
    \begin{small}
    \setlength{\tabcolsep}{4.5mm}{
    \begin{tabular}{c|ccc}
        \toprule
        \multirow{2}{*}{Problem Type} & \multicolumn{3}{c}{Number of Binary Variables}\\
        & \makebox[0.1\linewidth][c]{Task \uppercase\expandafter{\romannumeral1}} & \makebox[0.1\linewidth][c]{Task \uppercase\expandafter{\romannumeral2}} & \makebox[0.1\linewidth][c]{Task \uppercase\expandafter{\romannumeral3}}\\
        \midrule
        \makecell[c]{Problem \ref{problem:qcbo}\\(QCBO)} & 137 & 303 & 351\\
        \hline
        \makecell[c]{Problem \ref{problem:QUBO}\\(QUBO)} & 183 & 431 & 530\\
        \bottomrule
    \end{tabular}}
    \end{small}
    \end{center}
\end{table}

\begin{figure}[htbp]%
    \centering
     \includegraphics[width=0.43\textwidth]{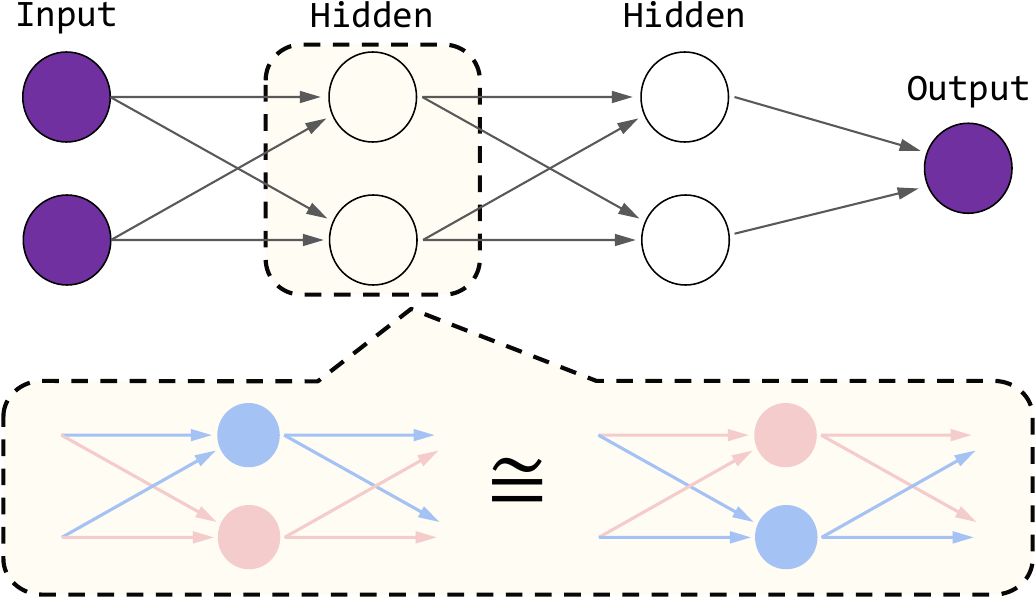}
    \caption{\textbf{Parameters after neuron exchange are viewed as identical parameters.}
             Any two neurons in the hidden layer could exchange.
             In this example, the blue neuron exchanges with the red one,
             resulting in different values of network parameters.
             To eliminate the effect,
             the network parameters after neuron exchange are viewed as the same.
             }
    \label{fig:neuron_exchange}
\end{figure}

We define two network parameters identical when every corresponding value in them is the same.
Because any two neurons in the hidden layer could exchange,
the parameters after neuron exchange are also considered as identical,
which is described in Figure \ref{fig:neuron_exchange}.
As shown in Table \ref{sol_identity},
the solved network parameters are completely identical,
implying the correctness of conversion process in Section \ref{sec:convert}.

\begin{table*}[htbp]
    \vskip -0.15in
    \caption{\textbf{The loss and accuracy in each task.}
             In all tasks, both the final losses and training accuracies for QCBO and QUBO have the same value.}
    \label{tab:indicator}
    \vskip 0.15in
    \begin{center}
    \begin{small}
    \begin{tabular}{c|c|ccc}
        \toprule
        \multirow{2}{*}{Problem Type} & \multirow{2}{*}{Indicator} & \multicolumn{3}{c}{Value}\\
         & & Task \uppercase\expandafter{\romannumeral1} & Task \uppercase\expandafter{\romannumeral2} & Task \uppercase\expandafter{\romannumeral3}\\
        \midrule
        \multirow{2}{*}{\makecell[c]{Problem \ref{problem:qcbo}\\(QCBO)}} & $\mathcal{L}_{\rm MSE}$ & 0.67 & 0 & 0\\
            & $\rm{accuracy}$ & 83.3\% & 100\% & 100\%\\
        \hline
        \multirow{2}{*}{\makecell[c]{Problem \ref{problem:QUBO}\\(QUBO)}} & $\mathcal{L}_{\rm QUBO}$ & 0.67 & 0 & 0\\
        & $\rm{accuracy}$ & 83.3\% & 100\% & 100\%\\
        \bottomrule
    \end{tabular}
    \end{small}
    \end{center}
\end{table*}

\begin{table}[htbp]
    \setlength{\tabcolsep}{0.04in}
    \caption{\textbf{The result of solution identity.}
             In each task, the solved parameters of QCBO and QUBO are identical.
             The values in bracket show the identity possibilities among 100 runs.
             The result implies the correctness of conversion process in Section \ref{sec:convert}.}
    \label{sol_identity}
    \vskip 0.15in
    \begin{center}
    \begin{small}
    \begin{tabular}{c|ccc}
        \toprule
        \multirow{2}{*}{Question} & \multicolumn{3}{c}{Answer}\\
        & Task \uppercase\expandafter{\romannumeral1} & Task \uppercase\expandafter{\romannumeral2} & Task \uppercase\expandafter{\romannumeral3}\\
        \midrule
        \multirow{2}{*}{\makecell[c]{Are the solved network parameters\\$\theta^*_{\rm QCBO}$ and $\theta^*_{\rm QUBO}$ identical?}} & Yes & Yes & Yes\\
        & (100\%)& (100\%)& (100\%)\\
        \bottomrule
    \end{tabular}
    \end{small}
    \end{center}
\end{table}

\subsection{Verification of Problem Solvability}
\label{verification:solvability}
\begin{figure}[htbp]%
    \centering
    \includegraphics[width=0.43\textwidth]{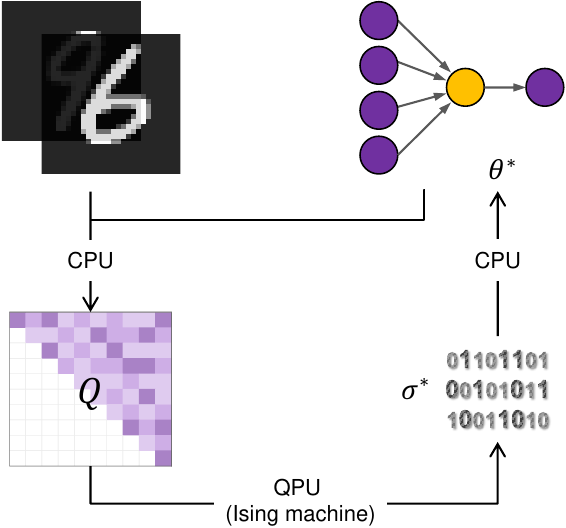}
    \caption{\textbf{Training workflow on MNIST with QPU and CPU.}
             The coefficient matrix $Q$ in QUBO problem is calculated on CPU based on the given dataset and network structure.
             The optimal solution $\sigma^*$ is solved on QPU, i.e., Ising machines.
             The optimal network parameter $\theta^*$ is decoded from $\sigma^*$ using CPU.
             }
    \label{fig:MNIST_CPU_QPU}
\end{figure}

To verify the solvability of Problem \ref{problem:QUBO} (QUBO),
we implement an experiment on simplified MNIST handwritten-digit dataset with a simulated Ising machine.
The overall training workflow is shown in Figure \ref{fig:MNIST_CPU_QPU}.
MNIST is a classical dataset containing images of 10 handwritten-digits \cite{lecun1998mnist},
and its simplified version is widely used for validating quantum neural networks \cite{jiang2021co, carrasquilla2023quantum}.
Two digits, 6 and 9, are selected in our paper for constructing a binary classification task.
We preprocess the images by dividing image into four patches and downsampling into $2\times 2$ pixels.
The pixel values are either $-1,0$ or $+1$, depending on the number of white pixels in their corresponding patches.
The preprocessed images are shown in Figure \ref{fig:MNIST}(a).
More preprocessing details can be found in Appendix \ref{appendix:MNIST}.
After preprocessing, only 4 images,
including two images of digit 6 and two images of digit 9,
are selected as the training dataset.
For the neural network, a QNN with 1 hidden layer and 1 hidden unit is used.
There are 4 units in its input layer,
where each unit receives one pixel value in image.
The output value in the last layer ranges in $[-1,+1]$.
It predicts for digit 6 when the output is non-negative,
otherwise for digit 9.
The neural network structure is shown in Figure \ref{fig:MNIST}(a).
We conduct the experiment using Fixstars Amplify AE \cite{fixstars},
which is a GPU-simulated Ising machine.

The number of binary variables used in QUBO is 108.
We set the annealing time as 700 milliseconds.
As shown in the loss histogram in Figure \ref{fig:MNIST}(b),
the success probability $P_s$ of finding the optimal solution,
i.e. achieving zero loss,
is 72\%.
For another performance index,
TTS (time-to-solution) is defined as $T_{\rm com}\log(1-0.99)/\log(1-P_s)$ \cite{hamerly2019experimental, aramon2019physics},
where $T_{\rm com}$ is the computation time per trial.
In this test,
TTS is 2.53 seconds,
which means the optimal network parameter can be found with 99\% success probability when the annealing time is set as 2.53 seconds.
For the solved network, its classification accuracy on test dataset is 98.3\%,
where the test dataset contains 1967 images.
The confusion matrix on test dataset is shown in Figure \ref{fig:MNIST}(c).
All of the results demonstrate the converted QUBO problem is solvable on Ising machines.

\begin{figure*}[htbp]%
    \centering\hspace{-0.4in}
    \includegraphics[width=0.88\textwidth]{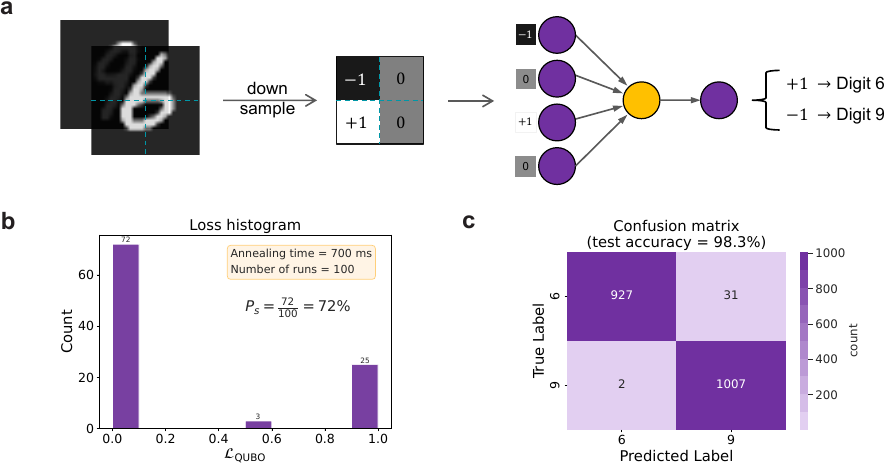}
    \caption{\textbf{Verification on simplified MNIST dataset using Fixstars Amplify AE.}
            (a) Dataset and network structure: Two digits, 6 and 9, are selected to construct a classification task.
            The images are downsampled to $2\times 2$ pixels.
            The network has 4 input neurons corresponding to the 4 pixels.
            There are 1 hidden neuron and 1 output neuron, where positive and negative outputs predict as digit 6 and digit 9, respectively.
            (b) Loss histogram: There are 72 runs achieving zero loss among 100 runs, implying the success probability $P_s$ of finding optimal solution is 72\%.
            (c) Confusion matrix: The classification accuracy on test dataset is 98.3\%,
            where the test dataset has 1967 images.
            The results successfully demonstrate the solvability of converted QUBO problem on Ising machines.
             }
    \label{fig:MNIST}
\end{figure*}

\section{Discussion}
\label{sec:discuss}

\subsection{Algorithm Applicability}

In this paper, we have successfully implemented fundamental network topology and learning paradigm in Ising learning algorithm, i.e. linear layer, \textit{sign} activation, and MSE loss.
The realizations of linear layer and \textit{sign} activation rely on the constraint representation in Section \ref{sec:QCBO}.
By tailoring those constraints and changing their types,
many other modules can be supported in our algorithm,
such as convolution layer,
pooling layer,
normalization layer,
\textit{Leaky ReLU} activation,
and hinge loss, etc.
All the supported modules are listed in Table \ref{tab:applicability},
and their constraint formulations are detailedly discussed in Appendix \ref{appendix:modules}.

More complex modules, e.g. attention layer and cross entropy loss,
are not supported yet.
It is because non-polynomial functions,
such as exponential, logarithmic, and sigmoid functions,
cannot be formulated in a polynomial constraint.
In order to match these modules,
one possible way is to use Taylor expansion,
where functions are approximated by few polynomial terms.
In this way,
more network structures may be realized, such as full precision network and Transformer network.

\begin{table}[H]
    \caption{\textbf{Applicability of Ising learning algorithm.}
             Supported modules are marked by checkmark,
             while unmarked modules need further research for adaptation.}
    \label{tab:applicability}
    \vskip 0.15in
    \begin{center}
    \begin{small}
    \begin{tabular}{c|c|c}
        \toprule
        \multicolumn{2}{c|}{Module} & Applicable\\
        \midrule
        \multirow{6}{*}{Layer} & Linear & \checkmark \\
              & Convolution & \checkmark \\
              & Pooling & \checkmark \\
              & Normalization & \checkmark \\
              & Recurrent & \\
              & Attention & \\
        \hline
        \multirow{8}{*}{Activation} & Sign & \checkmark \\
                   & ReLU & \checkmark \\
                   & Leaky ReLU & \checkmark \\
                   & PReLU & \checkmark \\
                   & Absolute & \checkmark\\
                   & Tanh & \\
                   & Sigmoid & \\
                   & ELU & \\
        \hline
        \multirow{3}{*}{Loss} & MSE & \checkmark \\
             & Hinge loss & \checkmark\\
             & Cross entropy &\\
        \bottomrule
    \end{tabular}
    \end{small}
    \end{center}
\end{table}

\subsection{Future Prospects}
Given Ising learning algorithm being the incipient approach to train multi-layer feedforward networks on Ising machines, it leaves ample room for future improvement.

On the one hand, more learning paradigms could be explored, e.g. self-supervised learning and reinforcement learning.
For example, the MSE loss can be replace by a max-margin contrastive loss,
which is a quadratic loss,
to implement self-supervised learning.
For reinforcement learning,
a quadratic loss can also be constructed based on self-consistent condition,
and Ising learning algorithm perhaps should be tailored as a iterative framework to adapt the shift of data distribution caused by consistent interaction between agent and environment.

On the other hand,
extra research is needed to reduce space complexity of Ising learning, in order to train larger network on larger dataset.
A possible way is using alternating direction method of multipliers (ADMM) 
to decouple the update of network parameter and sample-related variables.
As a result,
the sample-related variables no long need to be decision variables during optimizing stage of network parameter,
making the number of used spins independent to dataset size $N$.
Although the iteration framework of ADMM increases time complexity of Ising learning,
the space complexity is greatly reduced.

Moreover, the quantum approximate optimization algorithm (QAOA)
can be explored to solve the converted QUBO problem on universal quantum computers.
In this way,
QUBO problem becomes a general interface,
who builds a bridge between universal quantum computer and feedforward networks.

In conclusion,
as the hardware of Ising machine continues to develop, there is high potential for our Ising learning algorithm to be applied to large-scale network training.


\bibliography{biblio}

\begin{thebibliography}{41}
\providecommand{\natexlab}[1]{#1}
\providecommand{\url}[1]{\texttt{#1}}
\expandafter\ifx\csname urlstyle\endcsname\relax
  \providecommand{\doi}[1]{doi: #1}\else
  \providecommand{\doi}{doi: \begingroup \urlstyle{rm}\Url}\fi

\bibitem[Krizhevsky et~al.(2012)Krizhevsky, Sutskever, and
  Hinton]{krizhevsky2012imagenet}
Alex Krizhevsky, Ilya Sutskever, and Geoffrey~E Hinton.
\newblock Imagenet classification with deep convolutional neural networks.
\newblock \emph{Advances in neural information processing systems}, 25, 2012.

\bibitem[He et~al.(2016)He, Zhang, Ren, and Sun]{he2016deep}
Kaiming He, Xiangyu Zhang, Shaoqing Ren, and Jian Sun.
\newblock Deep residual learning for image recognition.
\newblock In \emph{Proceedings of the IEEE conference on computer vision and
  pattern recognition}, pages 770--778, 2016.

\bibitem[Huang et~al.(2017)Huang, Liu, Van Der~Maaten, and
  Weinberger]{huang2017densely}
Gao Huang, Zhuang Liu, Laurens Van Der~Maaten, and Kilian~Q Weinberger.
\newblock Densely connected convolutional networks.
\newblock In \emph{Proceedings of the IEEE conference on computer vision and
  pattern recognition}, pages 4700--4708, July 2017.

\bibitem[Vaswani et~al.(2017)Vaswani, Shazeer, Parmar, Uszkoreit, Jones, Gomez,
  Kaiser, and Polosukhin]{vaswani2017attention}
Ashish Vaswani, Noam Shazeer, Niki Parmar, Jakob Uszkoreit, Llion Jones,
  Aidan~N Gomez, {\L}ukasz Kaiser, and Illia Polosukhin.
\newblock Attention is all you need.
\newblock \emph{Advances in neural information processing systems}, 30, 2017.

\bibitem[Marandi et~al.(2014)Marandi, Wang, Takata, Byer, and
  Yamamoto]{marandi2014network}
Alireza Marandi, Zhe Wang, Kenta Takata, Robert~L Byer, and Yoshihisa Yamamoto.
\newblock Network of time-multiplexed optical parametric oscillators as a
  coherent ising machine.
\newblock \emph{Nature Photonics}, 8\penalty0 (12):\penalty0 937--942, 2014.

\bibitem[Inagaki et~al.(2016)Inagaki, Inaba, Hamerly, Inoue, Yamamoto, and
  Takesue]{inagaki2016large}
Takahiro Inagaki, Kensuke Inaba, Ryan Hamerly, Kyo Inoue, Yoshihisa Yamamoto,
  and Hiroki Takesue.
\newblock Large-scale ising spin network based on degenerate optical parametric
  oscillators.
\newblock \emph{Nature Photonics}, 10\penalty0 (6):\penalty0 415--419, 2016.

\bibitem[Neukart et~al.(2017)Neukart, Compostella, Seidel, Von~Dollen, Yarkoni,
  and Parney]{neukart2017traffic}
Florian Neukart, Gabriele Compostella, Christian Seidel, David Von~Dollen,
  Sheir Yarkoni, and Bob Parney.
\newblock Traffic flow optimization using a quantum annealer.
\newblock \emph{Frontiers in ICT}, 4:\penalty0 29, 2017.

\bibitem[King et~al.(2021)King, Raymond, Lanting, Isakov, Mohseni,
  Poulin-Lamarre, Ejtemaee, Bernoudy, Ozfidan, Smirnov,
  et~al.]{king2021scaling}
Andrew~D King, Jack Raymond, Trevor Lanting, Sergei~V Isakov, Masoud Mohseni,
  Gabriel Poulin-Lamarre, Sara Ejtemaee, William Bernoudy, Isil Ozfidan,
  Anatoly~Yu Smirnov, et~al.
\newblock Scaling advantage over path-integral monte carlo in quantum
  simulation of geometrically frustrated magnets.
\newblock \emph{Nature communications}, 12\penalty0 (1):\penalty0 1113, 2021.

\bibitem[Barahona(1982)]{barahona1982computational}
Francisco Barahona.
\newblock On the computational complexity of ising spin glass models.
\newblock \emph{Journal of Physics A: Mathematical and General}, 15\penalty0
  (10):\penalty0 3241, 1982.

\bibitem[Lucas(2014)]{lucas2014ising}
Andrew Lucas.
\newblock Ising formulations of many np problems.
\newblock \emph{Frontiers in physics}, 2:\penalty0 5, 2014.

\bibitem[Cen et~al.(2022)Cen, Ding, Hao, Guan, Qin, Lyu, Li, Zhu, Xu, Dai,
  et~al.]{cen2022large}
Qizhuang Cen, Hao Ding, Tengfei Hao, Shanhong Guan, Zhiqiang Qin, Jiaming Lyu,
  Wei Li, Ninghua Zhu, Kun Xu, Yitang Dai, et~al.
\newblock Large-scale coherent ising machine based on optoelectronic parametric
  oscillator.
\newblock \emph{Light: Science \& Applications}, 11\penalty0 (1):\penalty0 333,
  2022.

\bibitem[Honjo et~al.(2021)Honjo, Sonobe, Inaba, Inagaki, Ikuta, Yamada,
  Kazama, Enbutsu, Umeki, Kasahara, et~al.]{honjo2021100}
Toshimori Honjo, Tomohiro Sonobe, Kensuke Inaba, Takahiro Inagaki, Takuya
  Ikuta, Yasuhiro Yamada, Takushi Kazama, Koji Enbutsu, Takeshi Umeki, Ryoichi
  Kasahara, et~al.
\newblock 100,000-spin coherent ising machine.
\newblock \emph{Science advances}, 7\penalty0 (40):\penalty0 eabh0952, 2021.

\bibitem[Willsch et~al.(2020)Willsch, Willsch, De~Raedt, and
  Michielsen]{willsch2020support}
Dennis Willsch, Madita Willsch, Hans De~Raedt, and Kristel Michielsen.
\newblock Support vector machines on the d-wave quantum annealer.
\newblock \emph{Computer physics communications}, 248:\penalty0 107006, 2020.

\bibitem[Neven et~al.(2012)Neven, Denchev, Rose, and Macready]{neven2012qboost}
Hartmut Neven, Vasil~S Denchev, Geordie Rose, and William~G Macready.
\newblock Qboost: Large scale classifier training withadiabatic quantum
  optimization.
\newblock In \emph{Asian Conference on Machine Learning}, pages 333--348. PMLR,
  2012.

\bibitem[Kumar et~al.(2018)Kumar, Bass, Tomlin, and Dulny]{kumar2018quantum}
Vaibhaw Kumar, Gideon Bass, Casey Tomlin, and Joseph Dulny.
\newblock Quantum annealing for combinatorial clustering.
\newblock \emph{Quantum Information Processing}, 17:\penalty0 1--14, 2018.

\bibitem[Korenkevych et~al.(2016)Korenkevych, Xue, Bian, Chudak, Macready,
  Rolfe, and Andriyash]{korenkevych2016benchmarking}
Dmytro Korenkevych, Yanbo Xue, Zhengbing Bian, Fabian Chudak, William~G
  Macready, Jason Rolfe, and Evgeny Andriyash.
\newblock Benchmarking quantum hardware for training of fully visible boltzmann
  machines.
\newblock \emph{arXiv preprint arXiv:1611.04528}, 2016.

\bibitem[B{\"o}hm et~al.(2022)B{\"o}hm, Alonso-Urquijo, Verschaffelt, and
  Van~der Sande]{bohm2022noise}
Fabian B{\"o}hm, Diego Alonso-Urquijo, Guy Verschaffelt, and Guy Van~der Sande.
\newblock Noise-injected analog ising machines enable ultrafast statistical
  sampling and machine learning.
\newblock \emph{Nature Communications}, 13\penalty0 (1):\penalty0 5847, 2022.

\bibitem[Adachi and Henderson(2015)]{adachi2015application}
Steven~H Adachi and Maxwell~P Henderson.
\newblock Application of quantum annealing to training of deep neural networks.
\newblock \emph{arXiv preprint arXiv:1510.06356}, 2015.

\bibitem[Niazi et~al.(2023)Niazi, Aadit, Mohseni, Chowdhury, Qin, and
  Camsari]{niazi2023training}
Shaila Niazi, Navid~Anjum Aadit, Masoud Mohseni, Shuvro Chowdhury, Yao Qin, and
  Kerem~Y Camsari.
\newblock Training deep boltzmann networks with sparse ising machines.
\newblock \emph{arXiv preprint arXiv:2303.10728}, 2023.

\bibitem[Laydevant et~al.(2023)Laydevant, Markovic, and
  Grollier]{laydevant2023training}
J{\'e}r{\'e}mie Laydevant, Danijela Markovic, and Julie Grollier.
\newblock Training an ising machine with equilibrium propagation.
\newblock \emph{arXiv preprint arXiv:2305.18321}, 2023.

\bibitem[Laydevant et~al.(2021)Laydevant, Ernoult, Querlioz, and
  Grollier]{laydevant2021training}
J{\'e}r{\'e}mie Laydevant, Maxence Ernoult, Damien Querlioz, and Julie
  Grollier.
\newblock Training dynamical binary neural networks with equilibrium
  propagation.
\newblock In \emph{Proceedings of the IEEE/CVF conference on computer vision
  and pattern recognition}, pages 4640--4649, 2021.

\bibitem[Salakhutdinov and Larochelle(2010)]{salakhutdinov2010efficient}
Ruslan Salakhutdinov and Hugo Larochelle.
\newblock Efficient learning of deep boltzmann machines.
\newblock In \emph{Proceedings of the thirteenth international conference on
  artificial intelligence and statistics}, pages 693--700. JMLR Workshop and
  Conference Proceedings, 2010.

\bibitem[Date et~al.(2021)Date, Arthur, and Pusey-Nazzaro]{date2021qubo}
Prasanna Date, Davis Arthur, and Lauren Pusey-Nazzaro.
\newblock Qubo formulations for training machine learning models.
\newblock \emph{Scientific reports}, 11\penalty0 (1):\penalty0 10029, 2021.

\bibitem[Li(2023{\natexlab{a}})]{rlbook}
Shengbo~Eben Li.
\newblock \emph{Reinforcement learning for sequential decision and optimal
  control}.
\newblock Springer Verlag, Singapore, 2023{\natexlab{a}}.

\bibitem[Duan et~al.(2021)Duan, Guan, Li, Ren, Sun, and
  Cheng]{duan2021distributional}
Jingliang Duan, Yang Guan, Shengbo~Eben Li, Yangang Ren, Qi~Sun, and Bo~Cheng.
\newblock Distributional soft actor-critic: Off-policy reinforcement learning
  for addressing value estimation errors.
\newblock \emph{IEEE transactions on neural networks and learning systems},
  33\penalty0 (11):\penalty0 6584--6598, 2021.

\bibitem[Guan et~al.(2021)Guan, Li, Duan, Li, Ren, Sun, and
  Cheng]{guan2021direct}
Yang Guan, Shengbo~Eben Li, Jingliang Duan, Jie Li, Yangang Ren, Qi~Sun, and
  Bo~Cheng.
\newblock Direct and indirect reinforcement learning.
\newblock \emph{International Journal of Intelligent Systems}, 36\penalty0
  (8):\penalty0 4439--4467, 2021.

\bibitem[Li(2023{\natexlab{b}})]{li2023miscellaneous}
Shengbo~Eben Li.
\newblock Miscellaneous topics.
\newblock In \emph{Reinforcement Learning for Sequential Decision and Optimal
  Control}, pages 403--449. Springer, 2023{\natexlab{b}}.

\bibitem[Hubara et~al.(2016)Hubara, Courbariaux, Soudry, El-Yaniv, and
  Bengio]{hubara2016binarized}
Itay Hubara, Matthieu Courbariaux, Daniel Soudry, Ran El-Yaniv, and Yoshua
  Bengio.
\newblock Binarized neural networks.
\newblock \emph{Advances in neural information processing systems}, 29, 2016.

\bibitem[Bazaraa et~al.(2013)Bazaraa, Sherali, and
  Shetty]{bazaraa2013nonlinear}
Mokhtar~S Bazaraa, Hanif~D Sherali, and Chitharanjan~M Shetty.
\newblock \emph{Nonlinear programming: theory and algorithms}.
\newblock John wiley \& sons, 2013.

\bibitem[Zaman et~al.(2021)Zaman, Tanahashi, and Tanaka]{zaman2021pyqubo}
Mashiyat Zaman, Kotaro Tanahashi, and Shu Tanaka.
\newblock Pyqubo: Python library for mapping combinatorial optimization
  problems to qubo form.
\newblock \emph{IEEE Transactions on Computers}, 71\penalty0 (4):\penalty0
  838--850, 2021.

\bibitem[Mandal et~al.(2020)Mandal, Roy, Upadhyay, and
  Ushijima-Mwesigwa]{mandal2020compressed}
Avradip Mandal, Arnab Roy, Sarvagya Upadhyay, and Hayato Ushijima-Mwesigwa.
\newblock Compressed quadratization of higher order binary optimization
  problems.
\newblock In \emph{Proceedings of the 17th ACM International Conference on
  Computing Frontiers}, pages 126--131, 2020.

\bibitem[Van~Amersfoort et~al.(2020)Van~Amersfoort, Smith, Teh, and
  Gal]{pmlr-v119-van-amersfoort20a}
Joost Van~Amersfoort, Lewis Smith, Yee~Whye Teh, and Yarin Gal.
\newblock Uncertainty estimation using a single deep deterministic neural
  network.
\newblock In Hal~Daumé III and Aarti Singh, editors, \emph{Proceedings of the
  37th International Conference on Machine Learning}, volume 119 of
  \emph{Proceedings of Machine Learning Research}, pages 9690--9700. PMLR,
  13--18 Jul 2020.

\bibitem[Optimization(2008)]{gurobi}
Gurobi Optimization.
\newblock Gurobi optimizer, 2008.
\newblock \nolinkurl{https://www.gurobi.com/}.

\bibitem[Systems(2019)]{dwave}
D‑Wave Systems.
\newblock Tree decomposition solver, 2019.
\newblock
  \nolinkurl{https://docs.ocean.dwavesys.com/en/stable/docs\_samplers/README.html?highlight=tree\%20decomposition#tree-decomposition/}.

\bibitem[LeCun(1998)]{lecun1998mnist}
Yann LeCun.
\newblock The mnist database of handwritten digits, 1998.
\newblock \nolinkurl{http://yann. lecun. com/exdb/mnist/}.

\bibitem[Jiang et~al.(2021)Jiang, Xiong, and Shi]{jiang2021co}
Weiwen Jiang, Jinjun Xiong, and Yiyu Shi.
\newblock A co-design framework of neural networks and quantum circuits towards
  quantum advantage.
\newblock \emph{Nature communications}, 12\penalty0 (1):\penalty0 579, 2021.

\bibitem[Carrasquilla et~al.(2023)Carrasquilla, Hibat-Allah, Inack, Makhzani,
  Neklyudov, Taylor, and Torlai]{carrasquilla2023quantum}
Juan Carrasquilla, Mohamed Hibat-Allah, Estelle Inack, Alireza Makhzani, Kirill
  Neklyudov, Graham~W Taylor, and Giacomo Torlai.
\newblock Quantum hypernetworks: Training binary neural networks in quantum
  superposition.
\newblock \emph{arXiv preprint arXiv:2301.08292}, 2023.

\bibitem[Group(2021)]{fixstars}
Fixstars Group.
\newblock Fixstars amplify annealing engine, 2021.
\newblock \nolinkurl{https://amplify.fixstars.com/en/}.

\bibitem[Hamerly et~al.(2019)Hamerly, Inagaki, McMahon, Venturelli, Marandi,
  Onodera, Ng, Langrock, Inaba, Honjo, et~al.]{hamerly2019experimental}
Ryan Hamerly, Takahiro Inagaki, Peter~L McMahon, Davide Venturelli, Alireza
  Marandi, Tatsuhiro Onodera, Edwin Ng, Carsten Langrock, Kensuke Inaba,
  Toshimori Honjo, et~al.
\newblock Experimental investigation of performance differences between
  coherent ising machines and a quantum annealer.
\newblock \emph{Science advances}, 5\penalty0 (5):\penalty0 eaau0823, 2019.

\bibitem[Aramon et~al.(2019)Aramon, Rosenberg, Valiante, Miyazawa, Tamura, and
  Katzgraber]{aramon2019physics}
Maliheh Aramon, Gili Rosenberg, Elisabetta Valiante, Toshiyuki Miyazawa,
  Hirotaka Tamura, and Helmut~G Katzgraber.
\newblock Physics-inspired optimization for quadratic unconstrained problems
  using a digital annealer.
\newblock \emph{Frontiers in Physics}, 7:\penalty0 48, 2019.

\bibitem[Qin et~al.(2020)Qin, Gong, Liu, Bai, Song, and Sebe]{qin2020binary}
Haotong Qin, Ruihao Gong, Xianglong Liu, Xiao Bai, Jingkuan Song, and Nicu
  Sebe.
\newblock Binary neural networks: A survey.
\newblock \emph{Pattern Recognition}, 105:\penalty0 107281, 2020.

\end{thebibliography}
\bibliographystyle{unsrtnat}

\clearpage
\appendix
\onecolumngrid
\section{Example of Rosenberg Order Reduction}
\label{appendix:rosenberg}

In Section \ref{sec:convert}, Rosenberg Polynomial is used in order reduction method to convert high-order binary optimization problem into QUBO problem.
This appendix shows a simple example to illustrate how the order reduction method works.


\begin{example}[Order Reduction Process]
    Suppose there is an optimization problem $\sigma^* = \arg \min_\sigma \mathcal{L}$ with loss function $\mathcal{L}=\sigma_1\sigma_2\sigma_3+\sigma_1\sigma_2+\sigma_3$,
    where $\sigma = \{\sigma_1, \sigma_2, \sigma_3\}$ and $\sigma_1, \sigma_2, \sigma_3 \in \{0,1\}$.
    The loss $\mathcal{L}$ is a 3-order polynomial.
    We can reduce it to a 2-order loss by the following steps:
    
    Firstly, define an auxiliary variable $\sigma_4 \in \{0,1\}$, and replace $\sigma_1\sigma_2$ by $\sigma_4$ in $\mathcal{L}$, get $\mathcal{L}'=\sigma_4\sigma_3+\sigma_4+\sigma_3$.
    
    Secondly, construct a Rosenberg polynomial $h(\sigma_1,\sigma_2,\sigma_4)=3\sigma_4+\sigma_1 \sigma_2 - 2\sigma_1 \sigma_4 -2\sigma_2 \sigma_4$,
    then add it onto the loss with a positive large coefficient $\lambda$.
    The new loss becomes $\mathcal{L}'' = \mathcal{L}' + \lambda h$,
    which is a 2-order polynomial loss.
    The original optimization problem can be rewritten as
    \begin{align*}
        \sigma^* &= \arg \min_{\sigma, \sigma_4} \mathcal{L}'',
    \end{align*}
    where $\sigma = \{\sigma_1, \sigma_2, \sigma_3\}$ and $\sigma_1,\sigma_2,\sigma_3,\sigma_4 \in \{0,1\}$.
\end{example}

Any polynomial high-order binary optimization problem can be induced to quadratic order binary optimization
by iteratively running the steps in the above example.

\section{Configuration for Quantized Neural Network}
\label{appendix:configuration}

For simplicity,
we set some common configurations for QNNs that are used in our three verifications in Section \ref{sec:result}.

The weights $W^{(k)}$ are set as either $+1$ or $-1$ except the last layer to save the number of used spins.
Only some more essential parameters,
i.e. $W^{(L)}$, $b^{(1)}$, and $b^{(L)}$,
are set as quantized values with higher precision,
as these layers have more crucial roles in network's prediction \cite{qin2020binary}.

Additionally, the biases $b^{(k)}$ in middle layers ($k=2,3,...,L-1$) are frozen as $b^{(k)} = H-1$.
Such configuration ensures
$$s^{(k)} = W^{(k)}a^{(k-1)}-b^{(k)}\neq 0,\quad k=2,3,...,L-1,$$ making constraint (\ref{sign_3}) no longer needed in middle layers.
The above inequality is held because the element in $W^{(k)}a^{(k-1)}$ will never equal to $H-1$ under the condition that $W^{(k)}\in\{-1,1\}^{H\cdot H}$ and $a^{(k-1)}_i\in\{-1,1\}^{H}$.
In this way, the constraint for activation function in middle layers only need to involve one constraint (\ref{sign_1}),
rather than involving two constraints (\ref{sign_1}) and (\ref{sign_3}).
As a result, the auxiliary variable $t^{(k)}$ do not need to be defined in these layers,
which saves the number of used spins.

For the last layer, $W^{(L)}$ and $b^{(L)}$ are designed as quantized fractional values,
ensuring that predicted output $\hat{y}_i$ has standard value range.
The range is set as $\mathbb{Y}=[-1,1]$ and $\hat{y}_i \in \mathbb{Y}^m$.
The labels in dataset are also scaled into this range ahead of time.

Finally, the trainable parameters are $\theta = \{W^{(k)}|k=1,2,...,L\}\cup\{b^{(1)}, b^{(L)}\}$.
Under such configurations,
the constraint representations for all layers are shown in Appendix \ref{appendix:constraints} and the binary encoding representations for all variables are shown in Appendix \ref{appendix:decision_variable}.

\section{Constraints Representation for All Layers}
\label{appendix:constraints}

Under the configurations in Appendix \ref{appendix:configuration},
all constraints are listed layer-by-layer as follows.
\begin{itemize}
\item For the first layer,
the following linear constraint and activation constraint should be satisfied
\begin{align*}
    & {\small \rm (Linear\ constraint):}\quad W^{(1)} x_i + b^{(1)} = s^{(1)}_i,\ \forall i\\
    & {\small \rm (Activation\ constraint):}\ 
    \begin{cases}
        \ \ a^{(1)}_i \odot s^{(1)}_i= r^{(1)}_i,\ \forall i\\
        \ \ a^{(1)}_i + 2r^{(1)}_i = 1 + t^{(1)}_i,\ \forall i.
    \end{cases}
\end{align*}
The linear constraint represents the topology of linear transformation in feedforward.
The activation constraint represents the dynamic of activation function in feedforward.
Two types of equality constraints are involved for activation function, including (\ref{sign_1}) and (\ref{sign_3}).

\item For the $2,3,...,L-1$ layers,
the following linear constraint and activation constraint should be satisfied
\begin{align*}
    & {\rm (Linear\ constraint):}\quad W^{(k)} a^{(k-1)}_i + b^{(k)} = s^{(k)}_i,\ \forall i\\
    & {\rm (Activation\ constraint):}\quad a^{(k)}_i \odot s^{(k)}_i= r^{(k)}_i,\ \forall i.
\end{align*}
The activation constraint in these layers only involves one constraint (\ref{sign_1}),
rather than involving two constraints (\ref{sign_1}) and (\ref{sign_3}) like the first layer,
because we ensure $s^{(k)}_i \neq 0$ by fixing $b^{(k)} = H-1$.

\item For the last layer,
only the linear constraint should be satisfied
\begin{align*}
    & {\rm (Linear\ constraint):}\quad W^{(L)} a^{(L-1)}_i + b^{(L)} = \hat{y}_i,\ \forall i.
\end{align*}
\end{itemize}

\section{Binary Representation of Decision Variables}
\label{appendix:decision_variable}

All of the decision variables in Problem \ref{problem:qcbo} and \ref{problem:QUBO} are encoded by binary decision variables in $\sigma$.
Under the configurations in Appendix \ref{appendix:configuration},
detailed encoding expression is listed layer-by-layer as follows.

\begin{itemize}
\item For the first layer ($k=1$):

The decision variables include $W^{(1)},b^{(1)},s_i^{(1)},r^{(1)}_i,t^{(1)}_i,$ and $a^{(1)}_i,\ \forall i=1,2,...,N$.
They are encoded by binary variables $\sigma_w^{(1)}, \sigma_b^{(1)}, \sigma_s^{(1)}, \sigma_r^{(1)}, \sigma_t^{(1)},$ and $\sigma_a^{(1)}$, where
\begin{align*}
    & W^{(1)} = 2\cdot \sigma_w^{(1)} - 1\\
    & b^{(1)} = \sum_{j=0}^{\lfloor \log_2 n2^{B +1}\rfloor} 2^{j}\cdot \sigma_{b}^{(1)}(j)\\
    & s_i^{(1)} = \sum_{j=0}^{\lfloor \log_2 n2^{B+2} \rfloor} 2^{j}\cdot \sigma_s^{(1)}(i,j) - n2^{B},\ \forall i\\
    & r^{(1)}_i = \sum_{j=0}^{\lfloor \log_2 3n2^{B} \rfloor} 2^{j}\cdot \sigma^{(1)}_r(i,j),\ \forall i\\
    & t^{(1)}_i = \sum_{j=0}^{\lfloor \log_2 3n2^{B+1} \rfloor} 2^{j}\cdot \sigma^{(1)}_t(i,j),\ \forall i\\
    & a^{(1)}_i = 2\cdot \sigma_a^{(1)}(i) - 1,\ \forall i.
\end{align*}
The domains of binary variables are:
\begin{gather*}
    \sigma_w^{(1)}\in \{0,1\}^{H\cdot n}, \\
    \sigma_{b}^{(1)}(j),\ \sigma_s^{(1)}(i,j),\ \sigma^{(1)}_r(i,j),\ \sigma^{(1)}_t(i,j),\ \sigma^{(1)}_a(i) \in \{0,1\}^H,\ \forall i.
\end{gather*}

\item For the $2,3,...,L-1$ layers ($k=2,3,...,L-1$):

The decision variables include $W^{(k)},s_i^{(k)},r^{(k)}_i,$ and $a^{(k)}_i,\ \forall i=1,2,...,N$.
They are encoded by binary variables $\sigma_w^{(k)}, \sigma_s^{(k)}, \sigma_r^{(k)},$ and $\sigma_a^{(k)}$, where
\begin{align*}
    & W^{(k)} = 2\cdot \sigma_w^{(k)} - 1\\
    & s_i^{(k)} = \sum_{j=0}^{\lfloor \log_2 2H\rfloor} 2^{j}\cdot \sigma_s^{(k)}(i,j) - 1,\ \forall i\\
    & r^{(k)}_i = \sum_{j=0}^{\lfloor \log_2 2H \rfloor} 2^{j}\cdot \sigma^{(k)}_r(i,j),\ \forall i\\
    & a^{(k)}_i = 2\cdot \sigma_a^{(k)}(i) - 1,\ \forall i.
\end{align*}
The domains of binary variables are:
\begin{gather*}
    \sigma_w^{(k)}\in \{0,1\}^{H\cdot H}, \\
    \sigma_s^{(k)}(i,j),\ \sigma^{(k)}_r(i,j),\ \sigma^{(k)}_a(i) \in \{0,1\}^H,\ \forall i.
\end{gather*}

\item For the last layers ($k=L$):

The decision variables include $W^{(L)},b^{(L)},$ and $\hat{y}_i,\ \forall i=1,2,...,N$.
They are encoded by binary variables $\sigma_w^{(L)}, \sigma_b^{(L)},$ and $\sigma_y$, where
\begin{align*}
    & W^{(L)} = \frac{1}{H} \left( \sum_{j=0}^{\lfloor \log_2 2H\rfloor} 2^{j}\cdot \sigma_w^{(L)}(j) - H \right)\\
    & b^{(L)} = \frac{1}{H} \left( \sum_{j=0}^{\lfloor \log_2 2H\rfloor} 2^{j}\cdot \sigma_{b}^{(L)}(j) - H \right)\\
    & \hat{y}_i = \frac{1}{2H} \left( \sum_{j=0}^{\lfloor \log_2 4H\rfloor} 2^{j}\cdot \sigma_{y}(i,j) - 2H \right),\ \forall i.
\end{align*}
The domains of binary variables are:
\begin{gather*}
    \sigma_w^{(L)}(j)\in \{0,1\}^{m\cdot H}, \\
    \sigma_{b}^{(L)}(j),\ \sigma_{y}(i,j) \in \{0,1\}^m,\ \forall i.
\end{gather*}

\end{itemize}

\section{Topology Formulation for Diverse Modules}
\label{appendix:modules}

As shown in Table \ref{tab:applicability},
Ising learning algorithm has a wide application range,
which can also be extended further.
In this paper,
we have successfully implemented the fundamental network
topology and learning paradigm, i.e. the linear layer, \textit{sign} activation function, and MSE loss function.
There are many other modules,
which are marked by checkmark in Table \ref{tab:applicability},
can be realized by tailoring equality constraints.
Their constraint formulations are listed  as follows.

\subsection{Network Layer Formulation}
\subsubsection{Linear Layer}
To capture the topology of linear layer,
the following equality constraint should be satisfied
$$W^{(k)} a^{(k-1)} + b^{(k)} = s^{(k)},$$
where $W^{(k)}$ and $b^{(k)}$ are the weight and bias of the $k$-th layer,
$s^{(k)}$ is the pre-activation value in the $k$-th layer,
and $a^{(k-1)}$ is the post-activation value in the last layer.
\subsubsection{Convolution Layer}
To capture the topology of convolution layer,
the following equality constraint should be satisfied
$$\omega^{(k)} * a^{(k-1)} = s^{(k)},$$
where $*$ means the convolution operation,
$\omega^{(k)}$ is the convolution kernel of the $k$-th layer,
$s^{(k)}$ is the pre-activation value in the $k$-th layer,
and $a^{(k-1)}$ is the post-activation value in the last layer.

Specifically, for a 2D convolution layer with $3\times 3$ kernel,
the following equality constraint should be satisfied
\begin{align*}
    \sum_{i=0}^2 \sum_{j=0}^2 \omega^{(k)}(i,j)\ a^{(k-1)}(r+i,c+j) = s^{(k)}(r,c),\quad \forall r,c
\end{align*}
where $\omega^{(k)}$ is the $3\times 3$ convolution kernel of the $k$-th layer,
$s^{(k)}$ is the pre-activation value in the $k$-th layer,
$a^{(k-1)}$ is the post-activation value in the last layer,
and $r$ and $c$ means the row index and column index in 2D feature.
\subsubsection{Pooling Layer}
As for average pooling layer,
e.g. a 2D average pooling with $3\times3$ filter,
the following equality constraint should be satisfied to capture its topology:
$$ \frac{1}{9}\sum_{i=0}^2 \sum_{j=0}^2 a^{(k)}(r+i,c+j) = a_{\rm pooling}^{(k)}(r,c),\quad \forall r,c
$$
where $a^{(k)}$ is the post-activation value in the $k$-th layer,
$a_{\rm pooling}^{(k)}$ is the value after pooling operation,
and $r$ and $c$ means the row index and column index in 2D feature.

As for max pooling layer,
e.g. a 2D max pooling with $3\times3$ filter,
the following constraint should be satisfied to capture its topology:
\begin{align*}
    \sum_{i=0}^2 \sum_{j=0}^2 \omega^{(k)}(r,c,i,j)\ a^{(k)}(r+i&,c+j) = a_{\rm pooling}^{(k)}(r,c),\quad \forall r,c\\
    \sum_{i=0}^2 \sum_{j=0}^2 \omega^{(k)}(r,c,&i,j) = 1,\quad \forall r,c\\
    \omega^{(k)}(r,c,i,j)\ a^{(k)}(r+i,c+j) +
    \left(1-\omega^{(k)}(r,c,i,j)\right)&M \geq a^{(k)}(r+i',c+j'),\quad \forall r,c\ \forall i,j,i',j' \in \{0,1,2\}\\
    \omega^{(k)}(r,c,i,j) \in \{0,1\},&\quad \forall r,c\ \forall i,j \in \{0,1,2\}
\end{align*}
where $a^{(k)}$ is the post-activation value in the $k$-th layer,
$a_{\rm pooling}^{(k)}$ is the value after pooling operation,
$\omega^{(k)}$ is the binary indicator implying whether one value is the maximal value among a convolution window,
$M$ is a big enough positive value,
and $r$ and $c$ means the row index and column index in 2D feature.
To handle the above constraint more conveniently,
the inequality constraint can be converted to an equality constraint by introducing auxiliary variable like Section \ref{sec:QCBO} does.

\subsubsection{Normalization Layer}
Normalization layer includes batch normalization,
layer normalization, etc.
Here we talk about batch normalization as an example,
while the others can be realized in the same way.
To capture the topology of batch normalization layer,
the following equality constraint should be satisfied:
$$s^{(k)} - \mu^{(k)}= s^{(k)}_{\rm norm} \odot \sigma^{(k)},$$
where $\mu^{(k)}$ and $\sigma^{(k)}$ represent the mean and standard deviation in the $k$-th layer,
$s^{(k)}$ is the pre-activation value in the $k$-th layer,
and $s^{(k)}_{\rm norm}$ is the value after normalization.

\subsection{Activation Function Formulation}
\subsubsection{Sign}
The \textit{sign} activation function is
\begin{align*}
    {\textit{sign}}\left(x\right) = 
    \begin{cases}
        +1,\ \ x\geq 0\\
        -1,\ \ x< 0
    \end{cases}.
\end{align*}
To capture the topology behavior of \textit{sign} activation function,
the following two constraints should be satisfied
\begin{align}
    & a^{(k)} \odot s^{(k)} = r^{(k)}\label{appendix:sign_1}\\
    & a^{(k)} + 2r^{(k)} \geq 1,\label{appendix:sign_2}
\end{align}
where $a^{(k)} \in \{-1,+1\}^H$ is the post-activation value in the $k$-th layer,
$s^{(k)} \in \mathbb{Z}^H$ is the pre-activation value in the $k$-th layer,
the operation $\odot$ represents element-wise multiplication,
and $r^{(k)} \in \mathbb{N}^H$ is an auxiliary variable.
The constraint (\ref{appendix:sign_1}) guarantees $a^{(k)}$ and $s^{(k)}$ have the same sign,
because $r^{(k)}$ is non-negative.
The value of $r^{(k)}$ will be equal to the absolute value of $s^{(k)}$ under constraint (\ref{appendix:sign_1}).
The constraint (\ref{appendix:sign_2}) guarantees $a^{(k)}=+1$ when $s^{(k)}=0$.
In order to handle constraint (\ref{appendix:sign_2}) more conveniently,
an auxiliary variable $t^{(k)} \in \mathbb{N}^H$ can be introduced to transfer the inequality constraint (\ref{appendix:sign_2}) into equality constraint:
\begin{align*}
    a^{(k)} + 2r^{(k)} = 1 + t^{(k)}.
\end{align*}

\subsubsection{ReLU}
The \textit{ReLU} activation function is
\begin{align*}
    {\textit{ReLU}}\left(x\right) = 
    \begin{cases}
        x,\ \ x\geq 0\\
        0,\ \ x< 0
    \end{cases}.
\end{align*}
To capture the topology behavior of \textit{ReLU} activation function,
the following two constraints should be satisfied
\begin{align}
    \frac{1}{2}\left(r^{(k)} + s^{(k)}\right) &= a^{(k)}\label{relu_1}\\
    t^{(k)} \odot s^{(k)} =\ &r^{(k)},\label{relu_2}
\end{align}
where $a^{(k)} \in \mathbb{N}^H$ is the post-activation value in the $k$-th layer,
$s^{(k)} \in \mathbb{Z}^H$ is the pre-activation value in the $k$-th layer,
the operation $\odot$ represents element-wise multiplication,
and $r^{(k)} \in \mathbb{N}^H$ and $t^{(k)} \in \{-1,+1\}^H$ are auxiliary variables.
The constraint (\ref{relu_2}) guarantees $t^{(k)}$ and $s^{(k)}$ have the same sign,
because $r^{(k)}$ is non-negative.
The value of $r^{(k)}$ will be equal to the absolute value of $s^{(k)}$ under constraint (\ref{relu_2}).
The constraint (\ref{relu_1}) guarantees $a^{(k)}=0$ if $s^{(k)}\leq0$, otherwise $a^{(k)}=s^{(k)}$.

\subsubsection{Leaky ReLU}
The \textit{Leaky ReLU} activation function is
\begin{align*}
    {\textit{Leaky-ReLU}}\left(x\right) = 
    \begin{cases}
        x,\ \ x\geq 0\\
        \alpha x,\ \ x< 0
    \end{cases}.
\end{align*}
To capture the topology behavior of \textit{Leaky ReLU} activation function,
the following two constraints should be satisfied
\begin{align}
    \left( \frac{1-\alpha}{2} t^{(k)} + \frac{1+\alpha}{2}\right) \odot &\ s^{(k)} = a^{(k)}\label{leaky_relu_1}\\
    t^{(k)} \odot s^{(k)} =\ &r^{(k)},\label{leaky_relu_2}
\end{align}
where $a^{(k)} \in \mathbb{N}^H$ is the post-activation value in the $k$-th layer,
$s^{(k)} \in \mathbb{Z}^H$ is the pre-activation value in the $k$-th layer,
the operation $\odot$ represents element-wise multiplication,
and $r^{(k)} \in \mathbb{N}^H$ and $t^{(k)} \in \{-1,+1\}^H$ are auxiliary variables.
The constraint (\ref{leaky_relu_2}) guarantees $t^{(k)}$ and $s^{(k)}$ have the same sign,
because $r^{(k)}$ is non-negative.
The value of $r^{(k)}$ will be equal to the absolute value of $s^{(k)}$ under constraint (\ref{leaky_relu_2}).
The constraint (\ref{leaky_relu_1}) guarantees $a^{(k)}=\alpha s^{(k)}$ if $s^{(k)}\leq0$, otherwise $a^{(k)}=s^{(k)}$.

\subsubsection{PReLU}
The \textit{PReLU} activation function is
\begin{align*}
    {\textit{PReLU}}\left(x\right) = 
    \begin{cases}
        x,\ \ x\geq 0\\
        \alpha x,\ \ x< 0
    \end{cases}.
\end{align*}
The only difference between \textit{PReLU} and \textit{Leaky ReLU} is that
$\alpha$ in \textit{PReLU} is a learnable parameter
rather than a constant.
To capture the topology behavior of \textit{PReLU} activation function,
the following two constraints should be satisfied:
\begin{align}
    \left( \frac{1-\alpha}{2} t^{(k)} + \frac{1+\alpha}{2}\right) \odot &\ s^{(k)} = a^{(k)}\label{prelu_1}\\
    t^{(k)} \odot s^{(k)} =\ &r^{(k)},\label{prelu_2}
\end{align}
where $a^{(k)} \in \mathbb{N}^H$ is the post-activation value in the $k$-th layer,
$s^{(k)} \in \mathbb{Z}^H$ is the pre-activation value in the $k$-th layer,
$\alpha \in \mathbb{R}^+$ is the slop value,
the operation $\odot$ represents element-wise multiplication,
and $r^{(k)} \in \mathbb{N}^H$ and $t^{(k)} \in \{-1,+1\}^H$ are auxiliary variables.
Practically,
$\alpha$ should be approximated as a bounded decimal fraction number by binary variables.
The constraint (\ref{prelu_2}) guarantees $t^{(k)}$ and $s^{(k)}$ have the same sign,
because $r^{(k)}$ is non-negative.
The value of $r^{(k)}$ will be equal to the absolute value of $s^{(k)}$ under constraint (\ref{prelu_2}).
The constraint (\ref{prelu_1}) guarantees $a^{(k)}=\alpha s^{(k)}$ if $s^{(k)}\leq0$, otherwise $a^{(k)}=s^{(k)}$.

\subsubsection{Absolute}
The \textit{absolute} activation function is
\begin{align*}
    {\textit{abs}}\left(x\right) = 
    \begin{cases}
        x,\ \ x\geq 0\\
        -x,\ \ x< 0
    \end{cases}.
\end{align*}
To capture the topology behavior of \textit{abs} activation function,
the following two constraints should be satisfied:
\begin{align}
    r^{(k)} \odot s^{(k)} = a^{(k)},\label{abs_1}
\end{align}
where $a^{(k)} \in \mathbb{N}^H$ is the post-activation value in the $k$-th layer,
$s^{(k)} \in \mathbb{Z}^H$ is the pre-activation value in the $k$-th layer,
the operation $\odot$ represents element-wise multiplication,
and $r^{(k)} \in \{-1,+1\}^H$ is an auxiliary variable.
The constraint (\ref{abs_1}) guarantees $r^{(k)}$ and $s^{(k)}$ have the same sign,
because $a^{(k)}$ is non-negative.
The value of $a^{(k)}$ will be equal to the absolute value of $s^{(k)}$ under this constraint.

\subsection{Loss Function Formulation}
\subsubsection{MSE Loss}
The MSE loss function is
\begin{align*}
    \mathcal{L}_{\rm MSE} = \frac{1}{N} \sum_{i=1}^N (y_i - \hat{y}_i)^2,
\end{align*}
where $N$ is the dataset size,
$y_i$ is the label of the $i$-th sample,
and $\hat{y}_i$ is the predicted value of the $i$-th sample.
It is already a quadratic loss function,
therefore it can be directly used in Ising learning algorithm.
\subsubsection{Hinge Loss}
The hinge loss function is
\begin{align*}
    \mathcal{L}_{\rm hinge} = \frac{1}{N} \sum_{i=1}^N \max (0,1-y_i\hat{y}_i),
\end{align*}
where $N$ is the dataset size,
$y_i\in \{-1,+1\}$ is the label of the $i$-th sample,
and $\hat{y}_i\in \mathbb{R}$ is the predicted value of the $i$-th sample.
To capture the behavior of hinge loss function,
the following constraint should be satisfied:
\begin{align}
    t_i \left(1-y_i \hat{y}_i\right) =\ r_i,\label{hinge_2}
\end{align}
where $r_i \in \mathbb{R}^+$ and $t_i \in \{-1,+1\}$ are auxiliary variables corresponding to the $i$-th sample.
Then the hinge loss becomes
$$\mathcal{L}_{\rm hinge} = \frac{1}{2N} \sum_{i=1}^N \left(r_i + 1- y_i\hat{y}_i\right)$$
Practically,
$r_i$ should be approximated as bounded decimal fraction number by binary variables.
The constraint (\ref{hinge_2}) guarantees $t_i$ and $1- y_i\hat{y}_i$ have the same sign,
because $r_i$ is non-negative.
The value of $r_i$ will be equal to the absolute value of $1- y_i\hat{y}_i$ under constraint (\ref{hinge_2}).

\section{Image Preprocessing on MNIST}
\label{appendix:MNIST}

The simplified MNIST handwritten-digit dataset is widely used for validating quantum neural networks \cite{jiang2021co, carrasquilla2023quantum}.
Two digits, 6 and 9, are selected in our paper to construct a binary classification task.
We preprocess the images by downsampling images into $2\times 2$ pixels.
The pixel values are either $-1,0$ or $+1$, depending on the number of white pixels in their corresponding patches.

\begin{figure*}[htbp]%
    \centering\hspace{-0.3in}
    \includegraphics[width=0.9\textwidth]{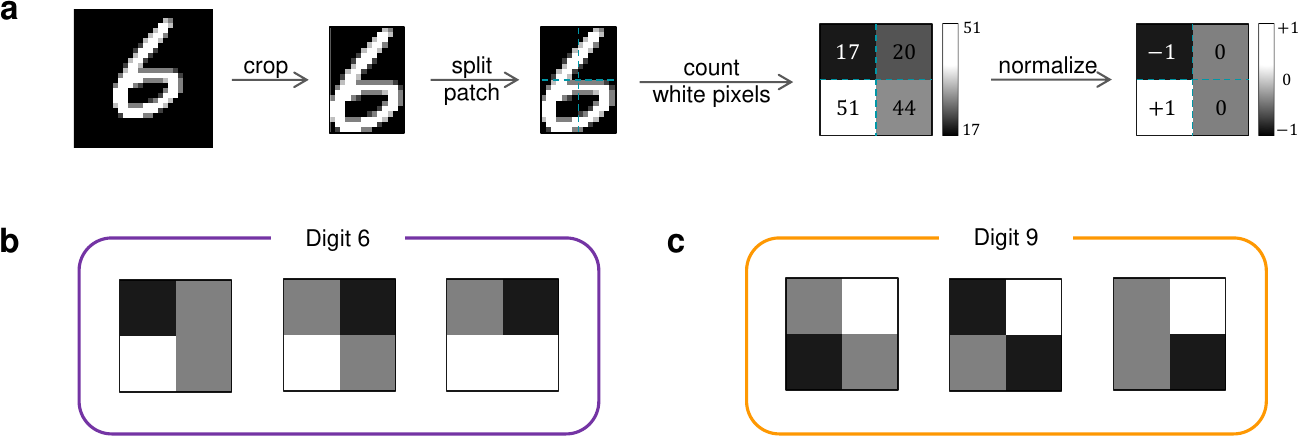}
    \caption{\textbf{Simplified MNIST dataset.}
             (a) Image preprocessing: The images are downsampled into $2\times 2$ pixels by cropping, splitting, counting and normalizing.
             (b) Image examples of digit 6.
             (c) Image examples of digit 9.
            }
    \label{fig:preprocess}
\end{figure*}

There are 4 steps in the preprocessing, including cropping, splitting, counting and normalizing.
In cropping step, the black margin is cropped out.
In splitting step, images are split into 4 patches.
In counting step, the number of white pixels in each patch is counted.
In normalizing step, the pixel values are normalized into either $-1,0$ or $+1$ depending on the number of white pixels in their conrresponding patches.
The preprocessing detail is shown in Figure \ref{fig:preprocess}(a).
The image examples of digit 6 and digit 9 are shown in Figure \ref{fig:preprocess}(b) and \ref{fig:preprocess}(c), respectively.

\end{document}